\documentclass[10pt,twocolumn,letterpaper]{article}

\usepackage{amsmath,amsfonts,bm}

\def\eqref#1{equation~\ref{#1}}

\def\1{\bm{1}}

\def\aa{\mathbf{a}}

\def\cc{\mathbf{c}}

\def\oo{\mathbf{o}}

\def\tt{\mathbf{t}}

\newcommand{\obs}{\oo}

\newcommand{\act}{\aa}
\newcommand\tuple[1]{\left\langle#1\right\rangle}
\newcommand{\cmd}{\cc}

\DeclareMathAlphabet{\mathsfit}{\encodingdefault}{\sfdefault}{m}{sl}
\SetMathAlphabet{\mathsfit}{bold}{\encodingdefault}{\sfdefault}{bx}{n}

\newcommand{\E}{\mathbb{E}}

\newcommand{\Var}{\mathrm{Var}}

\DeclareMathOperator*{\argmin}{arg\,min}

\usepackage{verbatim}
\usepackage{makecell}
\usepackage{iccv}
\usepackage{times}
\usepackage{epsfig}

\usepackage{url}
\usepackage{graphicx}

\usepackage{multirow}
\usepackage{amsmath,amssymb} 
\usepackage{color}
\usepackage{booktabs}
\usepackage[symbol]{footmisc}

\usepackage{blindtext}

\newcommand{\params}{\boldsymbol{\theta}}

\newcommand{\loss}{\ell}

\definecolor{dred}{rgb}{0.5,0.,0.}

\definecolor{dgreen}{rgb}{0.,0.5,0.}

\usepackage[pagebackref=true,breaklinks=true,letterpaper=true,colorlinks,bookmarks=false]{hyperref}

\iccvfinalcopy 

\ificcvfinal\fi
\begin{document}



\graphicspath{{./figures/}}

\title{
\vspace*{-12mm}
Exploring the Limitations of Behavior Cloning for Autonomous Driving}

\author{Felipe Codevilla  \footnotemark \\
Computer Vision Center (CVC)\\
Campus UAB, Barcelona, Spain\\
{\small fcodevilla@cvc.uab.es}
\and
Eder Santana\\
Toyota Research Institute (TRI)\\
Los Altos, CA, USA. \\
{\small edercsjr@gmail.com}
\and
Antonio M. L\'opez\\
Computer Vision Center (CVC)\\
Campus UAB, Barcelona, Spain\\
{\small alopez@cvc.uab.es}
\and
Adrien Gaidon\\
Toyota Research Institute (TRI)\\
Los Altos, CA, USA. \\
{\small adrien.gaidon@tri.global}
}

\begin{figure}

\twocolumn[{%
	\renewcommand\twocolumn[1][]{#1}%
	\maketitle
		\vspace{-4mm}
		\centering
 	\begin{tabular}{@{}c@{\hspace{1mm}}c@{\hspace{1mm}}c@{}}
		\includegraphics[width=0.33\linewidth]{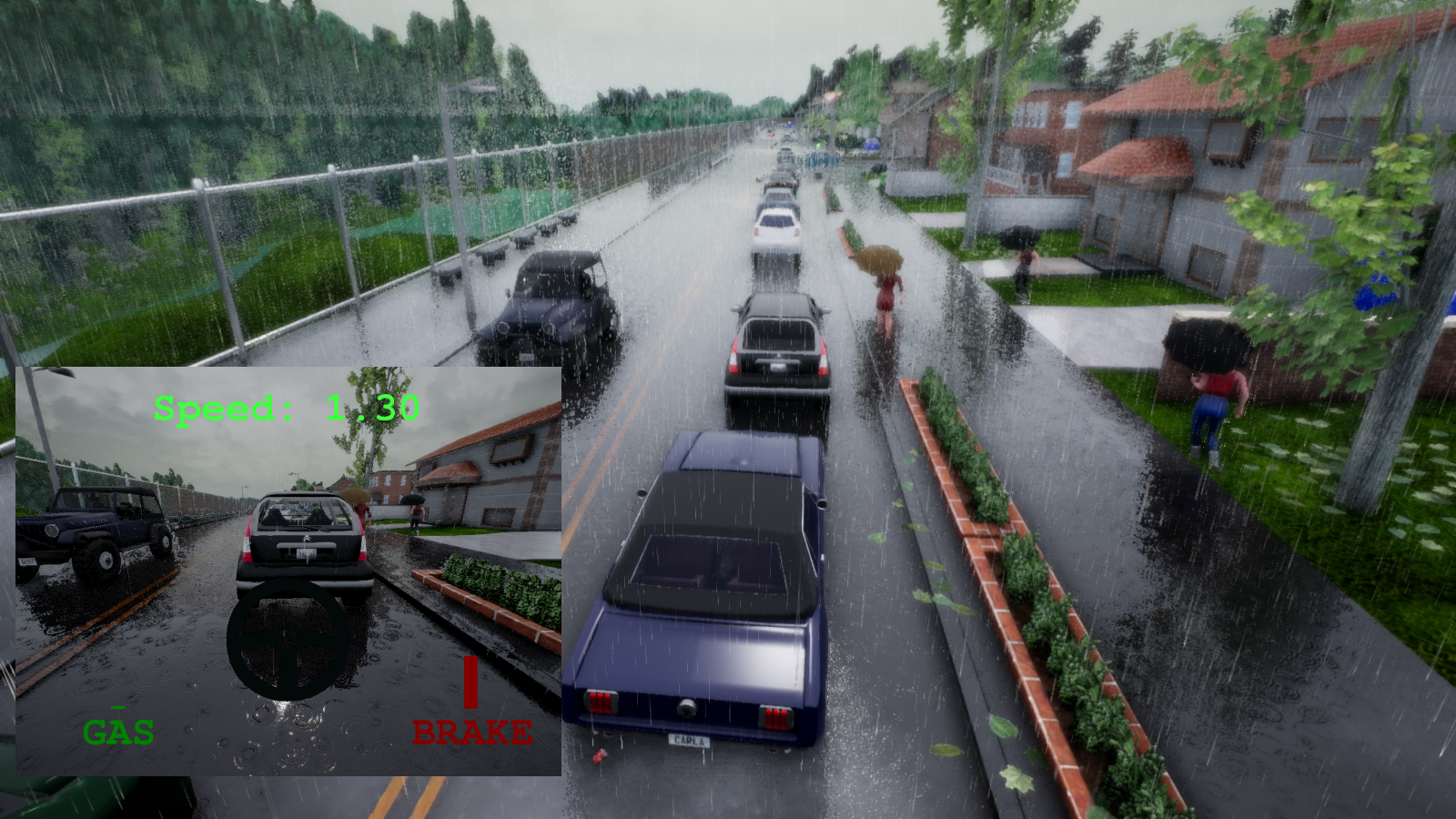} &
		\includegraphics[width=0.33\linewidth]{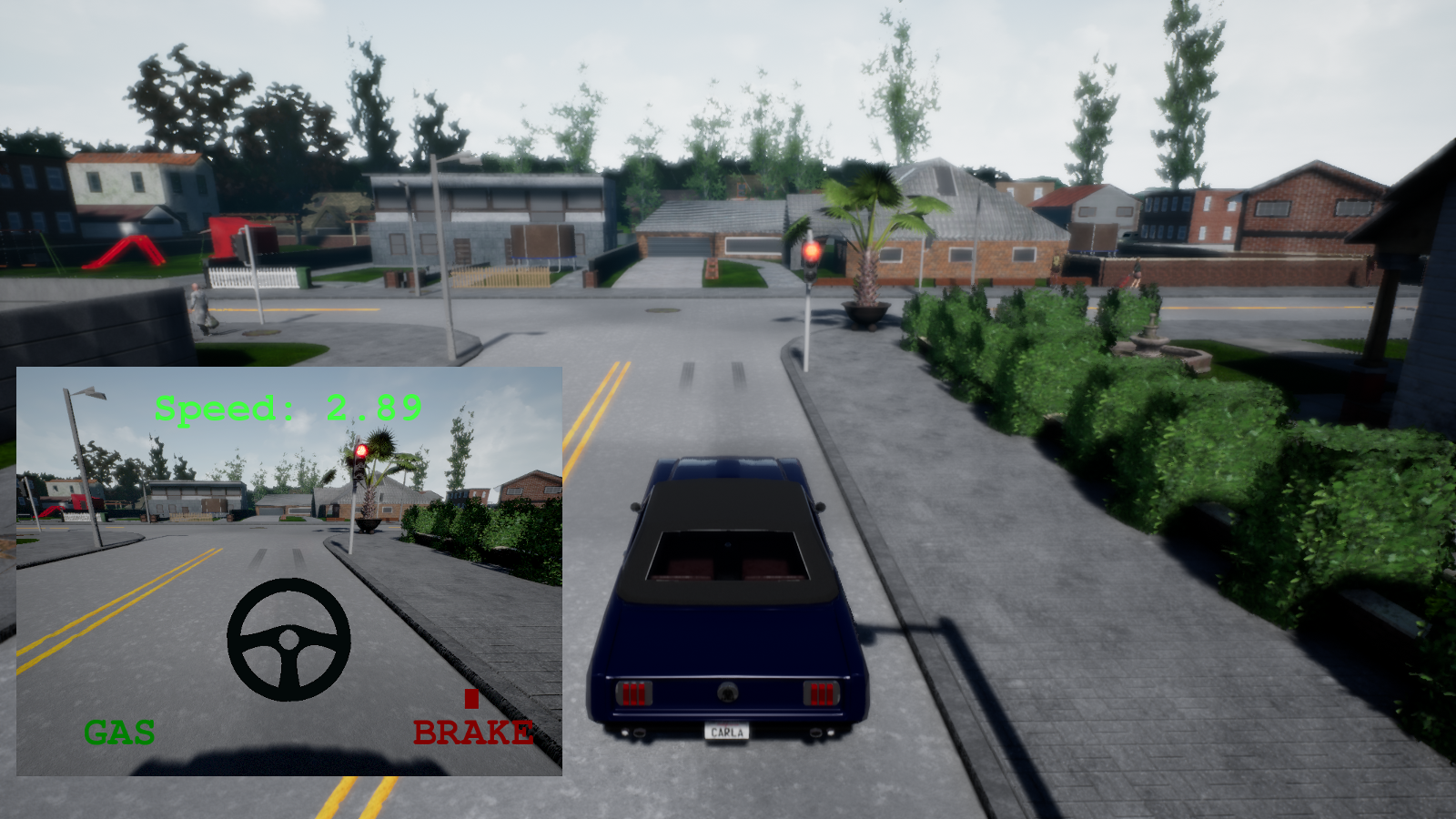} &
		\includegraphics[width=0.33\linewidth]{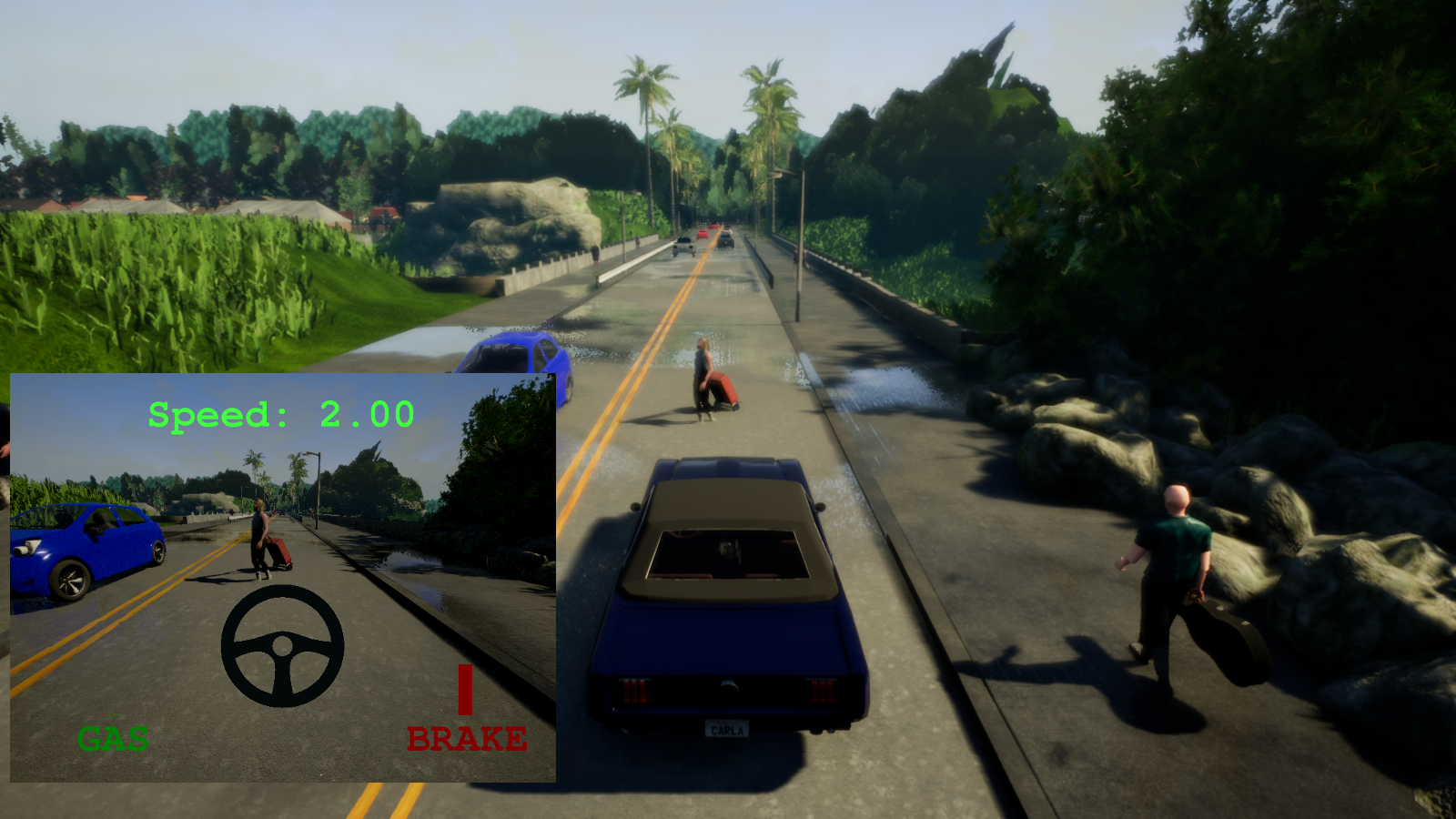} 
\vspace*{3mm}
\end{tabular}
		\caption{Driving scenarios from our new benchmark where the agent needs to react to dynamic changes in the environment, handle clutter (only part of the environment is causally relevant), and predict complex sensorimotor controls (lateral and longitudinal). We show that Behavior Cloning yields state-of-the-art policies in these complex scenarios and investigate its limitations.
		}
\vspace*{6mm}
\label{fig:teaser}
}]
\end{figure}


\vspace*{-10mm}
\begin{abstract}
\vspace*{-1mm}
Driving requires reacting to a wide variety of complex environment conditions and agent behaviors. Explicitly modeling each possible scenario is unrealistic. In contrast, imitation learning can, in theory, leverage data from large fleets of human-driven cars. Behavior cloning in particular has been successfully used to learn simple visuomotor policies end-to-end, but scaling to the full spectrum of driving behaviors remains an unsolved problem. 
In this paper, we propose a new benchmark to experimentally investigate the scalability and limitations of behavior cloning. 
We show that behavior cloning leads to state-of-the-art results, including in unseen environments, executing complex lateral and longitudinal maneuvers without these reactions being explicitly programmed.
However, we confirm well-known limitations (due to dataset bias and overfitting), new generalization issues (due to dynamic objects and the lack of a causal model), and training instability requiring further research before behavior cloning can graduate to real-world driving.
We will release our benchmark and code.

\footnotetext[1]{Work made during an internship at Toyota Research Institute (TRI)}

\end{abstract}

\section{Introduction}
\label{sec:introduction}
End-to-end behavior cloning for autonomous driving has recently attracted renewed interest \cite{chen2015deepdriving, Bojarski2016nvidiadriving, Codevilla2018, Wang18, muller2018driving} as a simple alternative to traditional modular approaches used in industry \cite{dickmanns2002development, leonard2008perception}. In this paradigm, perception and control are learned simultaneously using a deep network. Explicit sub-tasks are not defined, but may be implicitly learned from data. These sensorimotor controllers are typically obtained by imitation learning from human demonstrations~\cite{Abbeel2004inverserl, Ratcliff2007imitationlocomotion, Abbeel2006rlhelicopter, Silver2016}. The deep network learns, without being explicitly programmed, to recognize patterns associating sensory input (e.g., a single RGB image) with a desired reaction in terms of vehicle control parameters producing a target maneuver. 
Behavior cloning can directly learn from large fleets of human-driven vehicles without requiring a fixed ontology and extensive amounts of labeling. 
Finally, end-to-end imitative systems can be learned off-line in a safe way, in contrast to reinforcement learning approaches that typically require millions of trial and error runs in the target environment~\cite{levine2017learning} or a faithful simulation.


End-to-end imitative systems can suffer a domain shift between the off-line training experience and the on-line behavior \cite{Ross2011dagger}.
This problem, however, can be addressed in practice by data augmentation \cite{Bojarski2016nvidiadriving, Codevilla2018}.
Nonetheless, in spite of the early and recent successes of behavior cloning for end-to-end driving~\cite{Pomerleau1988, LeCun2005driving, chen2015deepdriving, Bojarski2016nvidiadriving, Codevilla2018}, it has not yet proved to scale to the full spectrum of driving behaviors, such as reacting to multiple dynamic objects.


In this paper, we propose a new benchmark, called \textit{NoCrash}, and perform a large scale analysis of end-to-end behavioral cloning systems in complex driving conditions not studied in this context before. We use a high fidelity simulated environment based on the open source CARLA simulator~\cite{Dosovitskiy2017} to enable reproducible large scale off-line training and on-line evaluation in over 80 hours of driving under several different conditions.
We describe a strong Conditional Imitation Learning baseline, derived from~\cite{Codevilla2018}, that significantly improves upon state of the art modular~\cite{Li18}, affordance based~\cite{Sauer2018conditional}, and reinforcement learning~\cite{Liang2018CIRL} approaches, both in terms of generalization performance in training environments and unseen ones.

Despite its positive performance, we identify limitations that prevent behavior cloning from successfully graduating to real-world applications.
First, although generalization performance should scale with training data, generalizing to complex conditions is still an open problem with a lot of room for improvement. In particular, we show that no approach reliably handles dense traffic scenes with many dynamic agents.
%
Second, we report generalization issues due to dataset biases and the lack of a causal model. We indeed observe diminishing returns after a certain amount of demonstrations, and even characterize a degradation of performance on unseen environments.
%
Third, we observe a significant variability in generalization performance when varying the initialization or the training sample order, similar to on-policy RL issues~\cite{henderson2018deep}. We conduct experiments estimating  the impact of ImageNet pre-training and show that it is not able to fully reduce the variance. This suggests the order of training samples matters for off-policy Imitation Learning, similar to the on-policy case~\cite{ZhangCho2017}.

Our paper is organized as follows. Section~\ref{sec:related} describes related work, Section~\ref{sec:behavior} our strong behavior cloning baseline, Section~\ref{sec:benchmark} our evalution protocol, including our new NoCrash benchmark, Section~\ref{sec:indicators} our experimental results, and Section~\ref{sec:conclusion} our conclusion.

\section{Related Work}
\label{sec:related}


\textbf{Behavior cloning} for driving dates back to the work of Pomerleau~\cite{Pomerleau1988} on lane following, later followed by other approaches~\cite{LeCun2005driving}, including going beyond driving~\cite{Abbeel2006rlhelicopter, soundararaj2009autonomous}.
The distributional shift between the training and testing distributions is the main known limitation of this approach, which might require \textit{on-policy} data collection~\cite{ross2010efficient, Ross2011dagger}, obtained by the learning agent.
Nonetheless, recent works have proposed effective \textit{off-policy} solutions, for instance by expanding the space of image/action pairs either using  noise~\cite{Laskey2017noise, Codevilla2018}, extra sensors~\cite{Bojarski2016nvidiadriving}, or modularization~\cite{Sauer2018conditional, li2018rethinking}.
We show, however, that there are other limitations important to consider in complex driving scenarios, in particular dataset bias and high variance, which both harm scaling generalization performance with training data.

\textbf{Dataset bias} is a core problem of real-world machine learning applications~\cite{torralba2011unbiased, barocas2017fairness} that can have dramatic effects in a safety-critical application like autonomous driving. Imitation learning approaches are particularly sensitive to this issue, as the learning objective might be dominated by the main modes in the training data.
Going beyond the original CARLA benchmark~\cite{Dosovitskiy2017}, we use our new NoCrash benchmark to quantitatively assess the magnitude of this problem on generalization performance for more realistic and challenging driving behaviors.

\textbf{High variance} is a key problem in powerful deep neural networks, and we show that high performance behavior cloning models are particularly suffering from this. This problem is related to sensitivity to both initialization and sampling order~\cite{neal2018modern}, reproducibility issues in Reinforcement Learning~\cite{henderson2018deep, machado2018revisiting}, and the need to move beyond the i.i.d.~data assumption towards curriculum learning~\cite{Bengio2009curriculum} for sensorimotor control~\cite{ZhangCho2017, andrychowicz2017hindsight}.

\textbf{Driving benchmarks} fall in two main categories: off-line datasets, e.g.,~\cite{Geiger2012, Santana:2016, Xu2017, Hecker_2018_ECCV}, or on-line environments. We focus here on on-line benchmarks, as visuomotor models performing well in dataset-based evaluations do not necessarily translate to good driving policies~\cite{Codevilla2018offline}.
Driving is obviously a safety-critical robotic application. Consequently, for safety and to enable reproducibility, researchers focus on using photo-realistic simulation environments. In particular, the CARLA open-source driving simulator \cite{Dosovitskiy2017} is emerging as a standard platform for driving research, used in \cite{Codevilla2018, muller2018driving, Sauer2018conditional, Liang2018CIRL, li2018rethinking}. Note, however, that transferring policies from simulation to the real-world is an open problem~\cite{Lopez17} out of the scope of this paper, although recent works have shown encouraging results~\cite{muller2018driving, Yang_2018_ECCV}.

\section{A Strong Baseline for Behavior Cloning}
\label{sec:behavior}
In this section, we first describe the behavior cloning framework we use, its limitations, and a robustified baseline that tries to tackle these issues.

\subsection{Conditional Imitation Learning}

Behavior cloning~\cite{Pomerleau1988, Schaal2003imitation, Ross2011dagger, levine2017learning} is a form of supervised learning that can learn sensorimotor policies from off-line collected data. The only requirements are pairs of input sensory observations associated with expert actions.
We use an expanded formulation for self-driving cars called Conditional Imitation Learning, CIL~\cite{Codevilla2018}. It uses a high-level navigational command $\cc$ that disambiguates imitation around multiple types of intersections. Given an expert policy $\pi^*(x)$ with access to the environment state $x$, we can execute this policy to produce a dataset,
$D=\{\tuple{\obs_i, \cmd_i, \act_i}\}_{i=1}^N$,
where $\obs_i$ are sensor data observations, $\cmd_i$ are high-level commands (e.g., take the next right, left, or stay in lane) and $\act_i=\pi^*(x_i)$ are the resulting vehicle actions (low-level controls).
Observations $ \obs_i = \{i,v_{m}\}$ contain a single image $i$ and the ego car speed $v_{m}$ \cite{Codevilla2018} added for the system to properly react to dynamic objects on the road. Without the speed context, the model cannot learn if and when it should accelerate or brake to reach a desired speed or stop.

We want to learn a policy $\pi$ parametrized by $\params$ to produce similar actions to $\pi^*$ based only on observations $\obs$ and high-level commands $\cmd$.
The best parameters $\params^*$ are obtained by minimizing an imitation cost $\loss$:
\begin{equation} \label{eq:imitation_with_latent_conditional}
  \params^* = \argmin \limits_{\params} \sum_{i} \loss\big(\pi(\obs_i, \cmd_i; \params), \act_i\big).
\end{equation}

In order to evaluate the performance of the learned policy $\pi(\obs_i, \cmd_i; \params)$ on-line at test time, we assume access to a score function giving a numeric value expressing the performance of the policy $\pi$ on a given benchmark (cf. section~\ref{sec:benchmark}).

\subsection{Limitations}
\label{sec:limitations_description}

In addition to the distributional shift problem~\cite{Ross2011dagger}, behavior cloning presents some key limitations.

\paragraph{Bias in Naturalistic Driving Datasets.}
The appeal of behavior cloning lies in its simplicity and theoretical scalability, as it can indeed learn by imitation from large off-line collected demonstrations (e.g., using driving logs from manually driven production vehicles).
It is, however, susceptible to dataset biases like all learning methods.
This is exacerbated in the case of imitation learning of driving policies, as most of real-world driving consists in either a few simple behaviors or a heavy tail of complex reactions to rare events.
Consequently, this can result in performance degrading as more data is collected, because the diversity of the dataset does not grow fast enough compared to the main mode of demonstrations.
This phenomenon was not clearly measured before. Using our new \emph{NoCrash} benchmark (section~\ref{sec:benchmark}), we confirm it may happen in practice.

\paragraph{Causal Confusion.}
Related to dataset bias, end-to-end behavior cloning can suffer from causal confusion~\cite{levine18nips}: spurious correlations cannot be distinguished from true causes in observed training demonstration patterns unless an explicit causal model or on-policy demonstrations are used.
Our new \emph{NoCrash} benchmark confirms the theoretical observation and toy experiments of~\cite{levine18nips} in realistic driving conditions. In particular, we identify a typical failure mode due to a subtle dataset bias: the \textit{inertia problem}.
When the ego vehicle is stopped (e.g., at a red traffic light), the probability it stays static is indeed overwhelming in the training data. This creates a spurious correlation between low speed and no acceleration, inducing excessive stopping and difficult restarting in the imitative policy.
Although mediated perception approaches that explicitly model causal signals like traffic lights do not suffer from this theoretical limitation, they still under-perform end-to-end learning in unconstrained environments, because not all causes might be modeled (e.g., some potential obstacles) and errors at the perception layer (e.g., missed detections) are irrecoverable.

\paragraph{High variance.}
With a fixed off-policy training dataset, one would expect CIL to always learn the same policy in different runs of the training phase. However, the cost function is optimized via Stochastic Gradient Descent (SGD), which assumes the data is independent and identically distributed~\cite{bottou2008tradeoffs}.
When training a reactive policy on snapshots of longer human demonstrations included in the training data, the i.i.d. assumption does not hold. Consequently, we might observe a high sensitivity to the initialization and the order in which the samples are seen during training. We confirm this in our experiments, finding an overall high variance due to both initialization and sampling order, following the decomposition in~\cite{neal2018modern}:
\begin{equation}
    \Var(\pi) = \E_D \big[ Var_I(\pi|D) \big]
    + Var_D \big( \E_I[\pi|D] \big),
    \label{eq:variance}
\end{equation}
where $I$ denotes the randomness in initialization.
Because the policy $\pi$ is evaluated on-line in simulated environments, we evaluate in practice the variance of the score on the test benchmark, and report results when freezing the initialization and/or varying the sampling order for different training datasets $D$ (including of varying sizes).

\subsection{Model}
\label{sec:model}

\begin{figure}[t]
  \centering
  { \fontsize{8pt}{10pt}\selectfont
  \setlength{\tabcolsep}{4pt}

    \includegraphics[width=0.95\linewidth]{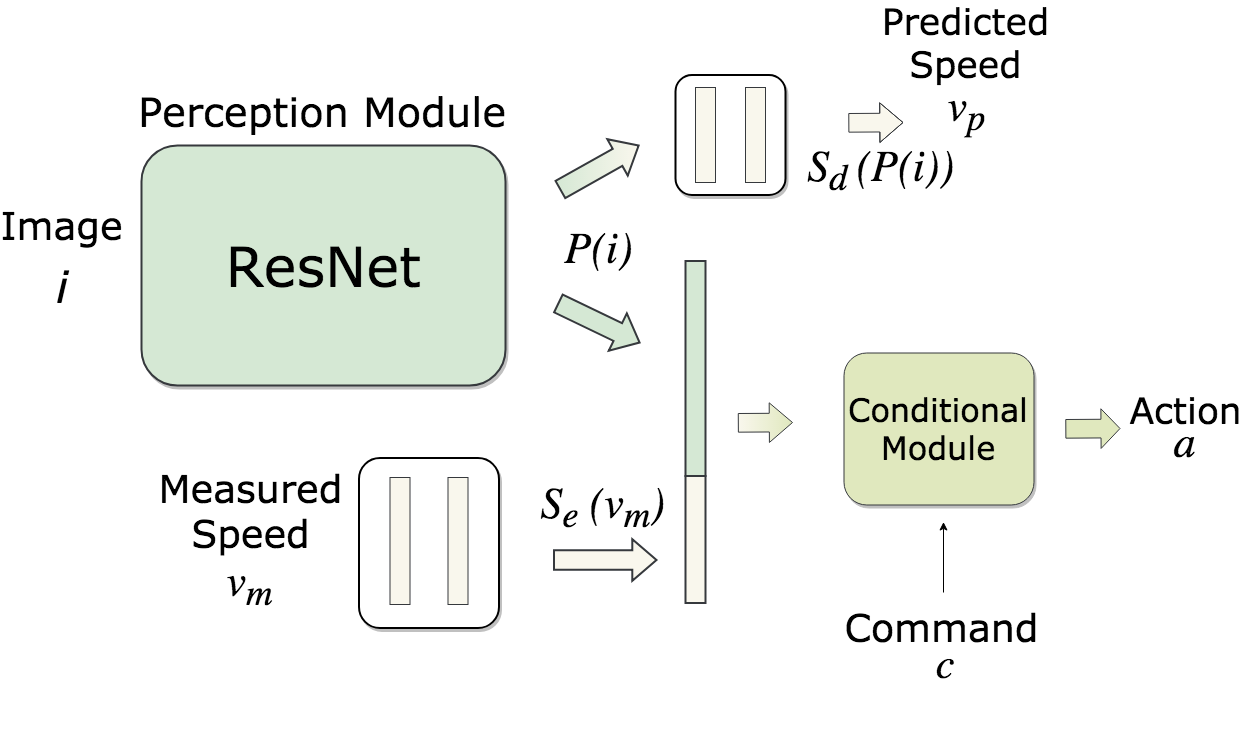} 
  }
  \caption{Our proposed network architecture, called CILRS, for end-to-end urban driving based on CIL~\cite{Codevilla2018}. A ResNet perception module processes an input image to a latent space followed by two prediction heads: one for controls and one for speed.}
  \label{fig:networkdiagram}
  \vspace{-1mm}
\end{figure}

In order to explore the aforementioned limitations of behavior cloning, we propose a robustified CIL model designed to improve on~\cite{Codevilla2018} while remaining strictly off-policy. Our network architecture, called CILRS, is shown in Figure~\ref{fig:networkdiagram}. We describe our enhancements below.

\paragraph{Deeper Residual Architecture.}
We use a ResNet34 architecture~\cite{he2016deep} for the perception backbone $\mathcal{P}(i)$.  In the presence of large amounts of data, using deeper architectures can be an effective strategy to improve performance~\cite{he2016deep}. In particular, it can reduce \emph{both} bias \emph{and} variance, maintaining in particular a constant variance due to training set sampling with both network width and depth~\cite{neal2018modern}.
For end-to-end driving, the choice of architecture has been mostly limited to small networks so far~\cite{Bojarski2016nvidiadriving, Codevilla2018, Sauer2018conditional} to avoid overfitting on limited datasets.
In contrast, we notice that bigger models have better generalization performance on learning reactions to dynamic objects and traffic lights in complex urban environments.

\paragraph{Speed Prediction Regularization.}
To cope with the inertia problem without an explicit mapping of potential causes or on-policy interventions, we jointly train a sensorimotor controller with a network that predicts the ego vehicle's speed. Both neural networks share the same representation via our ResNet perception backbone.  
Intuitively, what happens is that this joint optimization enforces the perception module to have speed related features into the learned representation. This reduces the dependency on input speed as the only way to get dynamics of the scene, leveraging instead visual cues that are predictive of the car's velocity (e.g., free space, curves, traffic light states, etc).

\paragraph{Other changes.}
We use L1 as loss function $\loss$ instead of the mean squared error (MSE), as it is more correlated to driving performance~\cite{Codevilla2018offline}.
As our \emph{NoCrash} benchmark consists of complex realistic driving conditions in the presence of dynamic agents, we collect demonstrations from an expert game AI using privileged information to drive correctly (i.e. always respecting rules of the road and not crashing into any obstacle). Robustness to heavy noise in the demonstrations is beyond the scope of our work, as we aim to explore limitations of behavior cloning methods \emph{in spite of} good demonstrations.
Finally, we pre-trained our perception backbone on ImageNet to reduce  initialization variance and benefit from generic transfer learning, a standard practice in deep learning seldom explored for behavior cloning.

\section{Evaluation}
\label{sec:benchmark}

In this section we discuss the simulated environment we use, CARLA, and review the original CARLA benchmark. Due to its limitations, we propose a new benchmark, called \textit{NoCrash}, that tries to better evaluate driving controllers reaction to dynamic objects. This new benchmark, thanks to its complexity, allows further analysis on limitations of behavior cloning and other policy learning methods.

\subsection{Simulated Environment}

We use the CARLA simulator~\cite{Dosovitskiy2017} version 0.8.4.
The CARLA environment is divided in two different towns. 
Town 01 contains 2.9 km of drivable roads in a suburban environment.
Town 02 is approximately 1.4 km of drivable roads, also in a suburban environment.

The CARLA environment may contain dynamic obstacles that interact with the ego car. Pedestrians, for instance, might cross the road on random occasions without any apparent previous notice. This action forces the ego car to promptly react.
The CARLA environment also contains a diversity of car brands that cruise at different speeds.
Overall it provides a diverse, photo-realistic, and dynamic environment with challenging driving conditions (cf. Figure~\ref{fig:teaser}).

The original CARLA benchmark \cite{Dosovitskiy2017} evaluates 
driving controllers on several goal directed tasks of increasing
difficulty. Three of the tasks consist of navigation
in an empty town and one of them in a town with
a small number of dynamic objects.
Each task is tested in four different conditions of increasingly different from the training environment. The conditions are: same as training, new
weather conditions that are derivatives from those seen during training, and a new town that has different buildings and different shadow patterns. Note that the biggest generalization test is the combination of new weather and new town.

The goal directed tasks are evaluated based on success rate. If the agent reaches the goal regardless of what happened during the episode, this episode is considered a success.
The collisions and other infractions are considered
and the average number of kilometers between infractions is measured.
This evaluation induces the benchmark to be mainly focused on problems of a static nature. These problems consider the environmental conditions and the static objects of the world like buildings and trees. Thus, the original CARLA benchmark mostly evaluates skills such as lane keeping and performing 90 degrees turns.

\subsection{NoCrash Benchmark}

We propose a new larger scale CARLA driving benchmark, called \emph{NoCrash}, designed to test the ability of ego vehicles to handle complex events caused by changing traffic conditions (e.g., traffic lights) and dynamic agents in the scene.
For this benchmark, we propose different tasks and metrics than the original CARLA benchmark~\cite{Dosovitskiy2017} to precisely measure specific reaction patterns that we know good drivers must master in urban conditions.
%


We propose three different tasks, each one corresponding to 25 goal directed episodes. In each episode, the agent starts at a random position and is directed by a high-level planner into reaching some goal position. The three tasks have the same set of start and end positions, as well as an increasing level of difficulty as follows:
\begin{enumerate}
    \item Empty Town: no dynamic objects.
    \item Regular Traffic: moderate number of cars and pedestrians.
    \item Dense Traffic: large number of pedestrians and heavy traffic (dense urban scenario).
\end{enumerate}

Similar to the CARLA Benchmark, \textit{NoCrash} has six
different weather conditions, where four were seen in training and two reserved for testing. It also has two different towns, one that is seen during training, and the other reserved for testing. For more details about the benchmark configuration, please refer to
the supplementary material.
As mentioned above, the measure of success of an episode should be more representative of the agent capabilities to react to dynamic objects.
The original CARLA benchmark \cite{Dosovitskiy2017} has a goal conditioned success rate metric that is computed separately from a kilometers between infractions metric.
The latter metric was proposed
to be analogous to the one commonly used by real-world driving 
evaluations where the number of human interventions per kilometer is counted \cite{kalra2016safety}. These interventions usually happen when the safety driver notices some inconsistent behavior that would lead the vehicle to
a possibly dangerous state. On a potentially inconsistent behavior, the human intervention will put the vehicle back to a safe state. However, in the CARLA benchmark analysis,
when an infraction is made, the episode continues after the infraction, leading to some inaccuracy in infraction counting. An example
of inaccuracy includes whether a crash after leaving the road be counted as one or two infractions.

In \textit{NoCrash}, instead of counting the number of infractions per kilometer, we end the episode as failing when any collision bigger
than a fixed magnitude happens.
With this limitation, we are setting a lower bound and have a guarantee of acceptable behaviors based on the measured percentage of success. Furthermore, this makes the evaluation even more similar to the km/interventions evaluation used in real world, since a new episode always sends the agent back to a safe starting state.  
In summary, we consider an episode to be successful if the agent reaches a certain goal under a time limit without colliding with any object.
We also care about the ability of the agent to obey traffic rules. In particular, we measure and report the percentage of traffic light violations in Supplementary material. Note that an episode is not terminated when a traffic light violation occurs unless they are followed by a collision.

\section{Experiments}
\label{sec:indicators}

\begin{table*}[t]
\small
\centering
\resizebox{1.0\linewidth}{!}{
 \begin{tabular}{@{}lcccccccccccccc@{}}
\toprule

       && \multicolumn{6}{c}{Training conditions} && \multicolumn{6}{c}{New town\,\&\,weather} \\
    Task             && CIL\cite{Codevilla2018}  & CIRL\cite{Liang2018CIRL}  & CAL\cite{Sauer2018conditional} & MT\cite{li2018rethinking} & CILR & CILRS && CIL\cite{Codevilla2018}  & CIRL\cite{Liang2018CIRL}  & CAL\cite{Sauer2018conditional} & MT\cite{li2018rethinking} & CILR & CILRS  \\
    \midrule
    Straight       && $98$ & $98$          & $\textbf{100}$ & $96$ & $94$ & $96$          && $80$  & $\textbf{98}$ & $94$ & $96$ & $92$          & $96$        \\
    One Turn       && $89$ & $\textbf{97}$ & $\textbf{97}$  & $87$ & $92$ & $92$          && $48$  & $80$          & $72$ & $82$ & $\textbf{92}$ & $\textbf{92}$\\
    Navigation     && $86$ & $93$          & $92$           & $81$ & $88$ & $\textbf{95}$ && $44$  & $68$          & $68$ & $78$ & $88$          & $\textbf{92}$\\
    Nav. Dynamic   && $83$ & $82$          & $83$           & $81$ & $85$ & $\textbf{92}$ && $42$  & $62$          & $64$ & $62$ & $82$          & $\textbf{90}$\\

\bottomrule
 \end{tabular}
}
\vspace{1mm}
\caption{Comparison with the state of the art on the original CARLA benchmark. The ``CILRS'' version corresponds to our CIL-based ResNet using the speed prediction branch, whereas ``CILR'' is without this speed prediction. These two models and CIL are the only ones that do not use any extra supervision or online interaction with the environment during training. The table reports the percentage of successfully completed episodes in each condition, selecting the best seed out of five runs.}
\label{tbl:comparison_legacy}
\vspace{1mm}
\end{table*}

\begin{table*}[t]
\small
\centering
\resizebox{1.0\linewidth}{!}{
 \begin{tabular}{@{}lcccccccccccccc@{}}
\toprule

       && \multicolumn{5}{c}{Training conditions} && \multicolumn{5}{c}{New Town\,\&\,Weather} \\
    Task             && CIL\cite{Codevilla2018}   & CAL\cite{Sauer2018conditional} & MT\cite{li2018rethinking} & CILR & CILRS  && CIL\cite{Codevilla2018}  & CAL\cite{Sauer2018conditional} & MT\cite{li2018rethinking} & CILR & CILRS   \\
    \midrule
Empty && $79 \pm 1$ &$81 \pm 1$ & $84 \pm 1$  &$92 \pm 1$ & $\textbf{97} \pm 2$ && $24 \pm 1$ &$25 \pm 3$ & $57 \pm 0$ & $66 \pm 2$ & $\textbf{90} \pm 2$ \\
Regular && $60 \pm 1$ &$73 \pm 2$ &$54 \pm 2$ &$72 \pm 5$ & $\textbf{83} \pm 0$ && $13 \pm 2$ &$14 \pm 2$ &$32 \pm 2$  &$54 \pm 2$ & $\textbf{56} \pm 2$\\
Dense && $21 \pm 2$ &$\textbf{42} \pm 3$ &$13 \pm 4$ & $28 \pm 1$ & $\textbf{42} \pm 2$ && $2 \pm 0$ &$10 \pm 0$ & $14 \pm 2$ & $13 \pm 4$ & $\textbf{24} \pm 8$ \\

\bottomrule
 \end{tabular}
}
\vspace{1mm}
\caption{Results on our \textit{NoCrash} benchmark. Mean and standard deviation on three runs, as CARLA 0.8.4 has significant non-determinism.}
\label{tbl:comparison_nocrash}
\vspace{1mm}
\end{table*}

In this section we detail our protocol for model training and briefly show that it is competitive with the state of the art.
We also explore several corner cases to explore
the limitations of the behavior cloning approach.




\subsection{Training Details}

First, we collected more than 400 hours of realistic simulated driving data from a single town of the CARLA environment using more than 200 GPU-days. We used an expert driving AI agent that leverages privileged information about the scene to drive naturally and well in complex conditions.
After automatically filtering the data for simulation failures, duplicates, and edge cases using simple rules, we built a dataset of 100 hours of driving, called CARLA100.
To enable running a wide range of experiments, we train all methods using a subset of 10 hours of expert demonstrations by default. We also report larger scale training experiments and scalability analyses in Section~\ref{sec:anlimitations} and in supplementary material.
We will release the code for our demonstrator and our CARLA100 training dataset for reproducibility. More details about them are given in the supplementary material.

Training controllers on this dataset, we found that augmentation was not as crucial as reported by previous works \cite{Codevilla2018, li2018rethinking}. The only regularization we found important for performance was using a $50\%$ dropout rate~\cite{srivastava2013dropout} after the last convolutional layer. Any larger dropout led us to under-fitting models.
All models were trained using Adam~\cite{Kingma2015adam} with minibatches  of $120$ samples and an initial learning rate of $0.0002$. At each iteration, a minibatch is sampled randomly from the entire dataset and presented to the network for training.
If we detect that the training error has not decreased for over
$1,000$ iterations we divide the learning rate by 10. 
We used a 2 hours validation dataset to discover when to stop the training process.
We validate every 20k iterations and if the validation error increases for three iterations we stop the training process and use this checkpoint to test on the benchmarks, both CARLA and \textit{NoCrash}.
We build a validation dataset as described in \cite{Codevilla2018offline}.

\subsection{Comparison with the state of the art}

We compare our results using both the original CARLA benchmark from \cite{Dosovitskiy2017} and our proposed \textit{NoCrash} benchmark. 
We compare two versions of our method: ``CILRS'' (our CIL extension with a ResNet architecture and speed prediction, as described in section~\ref{sec:behavior}), and a version without the speed prediction branch noted ``CILR''.
We compare our method with the original CIL from~\cite{Codevilla2018} and three state-of-the-art approaches: CAL~\cite{Sauer2018conditional}, MT~\cite{li2018rethinking}, and CIRL~\cite{Liang2018CIRL}. In contrast to end-to-end behavior cloning, these methods enforce some modularization that require extra information at training time, such as affordances (CAL), semantic segmentation (MT), or extra on-policy interaction with the environment (CIRL). Our approach only requires a fixed off-policy dataset of demonstrations.

We show results on the original CARLA benchmark~\cite{Dosovitskiy2017} in Table~\ref{tbl:comparison_legacy} and results on our proposed \emph{NoCrash} benchmark in Table~\ref{tbl:comparison_nocrash}.
While most methods perform well in most conditions on the original CARLA benchmark, they all perform significantly worse on \emph{NoCrash}, especially when trying to generalize to new conditions.
This confirms the usefulness of \emph{NoCrash} in terms of exploring the limitations of driving policy learning due to its more challenging nature. 

In addition, our proposed CILRS model significantly improves over the state of the art, e.g., $+9\%$ and $+26\%$ on CARLA ``Nav. Dynamic'' in training and new conditions respectively, $+10\%$ and $+24\%$ on \emph{NoCrash} Regular traffic in training and new conditions respectively.
The significant improvements in generalization conditions, both w.r.t. CIL and mediated approaches, confirm that our improved end-to-end behavior cloning architecture can effectively learn complex general policies from demonstrations alone.
Furthermore, our ablative analysis shows that speed prediction is helpful: CILR can indeed be up to $-14\%$ worse than CILRS on \emph{NoCrash}.

\subsection{Analysis of Limitations}
\label{sec:anlimitations}

Although clearly above the state of the art, our improved CILRS architecture nonetheless sees a strong degradation of performance similar to all other methods in the presence of challenging driving conditions.
We investigate how this degradation relates to the limitations of behavior cloning mentioned in Section~\ref{sec:limitations_description} by using the \textit{NoCrash} benchmark, in particular to better evaluate the interaction of the agents with dynamic objects.

\vspace*{-2mm}
\paragraph{Generalization in the presence of dynamic objects.}

Limited generalization was previously reported for end-to-end driving
approaches~\cite{Dosovitskiy2017}. In our experiments, we observed additional, and more prominent, generalization issues when the control policies have to deal with dynamic objects.
Table \ref{tbl:comparison_nocrash} indeed shows a large drop in performance as we change to tasks with more traffic, e.g., $-55\%$ and $-66\%$ from Empty to Dense traffic in \emph{NoCrash} training / new conditions respectively. In contrast, results in Empty town only degrade by $-7\%$ when changing to a new environment and weather. Therefore, the learned policies have a much harder time dealing robustly with a large number of vehicles and pedestrians.
Furthermore, this impacts all policy learning methods, including those using additional supervision or on-policy demonstrations, often even more than our proposed CILRS method.

\begin{figure*}
 \centering
  
  {
  \setlength{\tabcolsep}{2.7pt}
  \begin{tabular}{ccc}
    \includegraphics[height=0.26\linewidth]{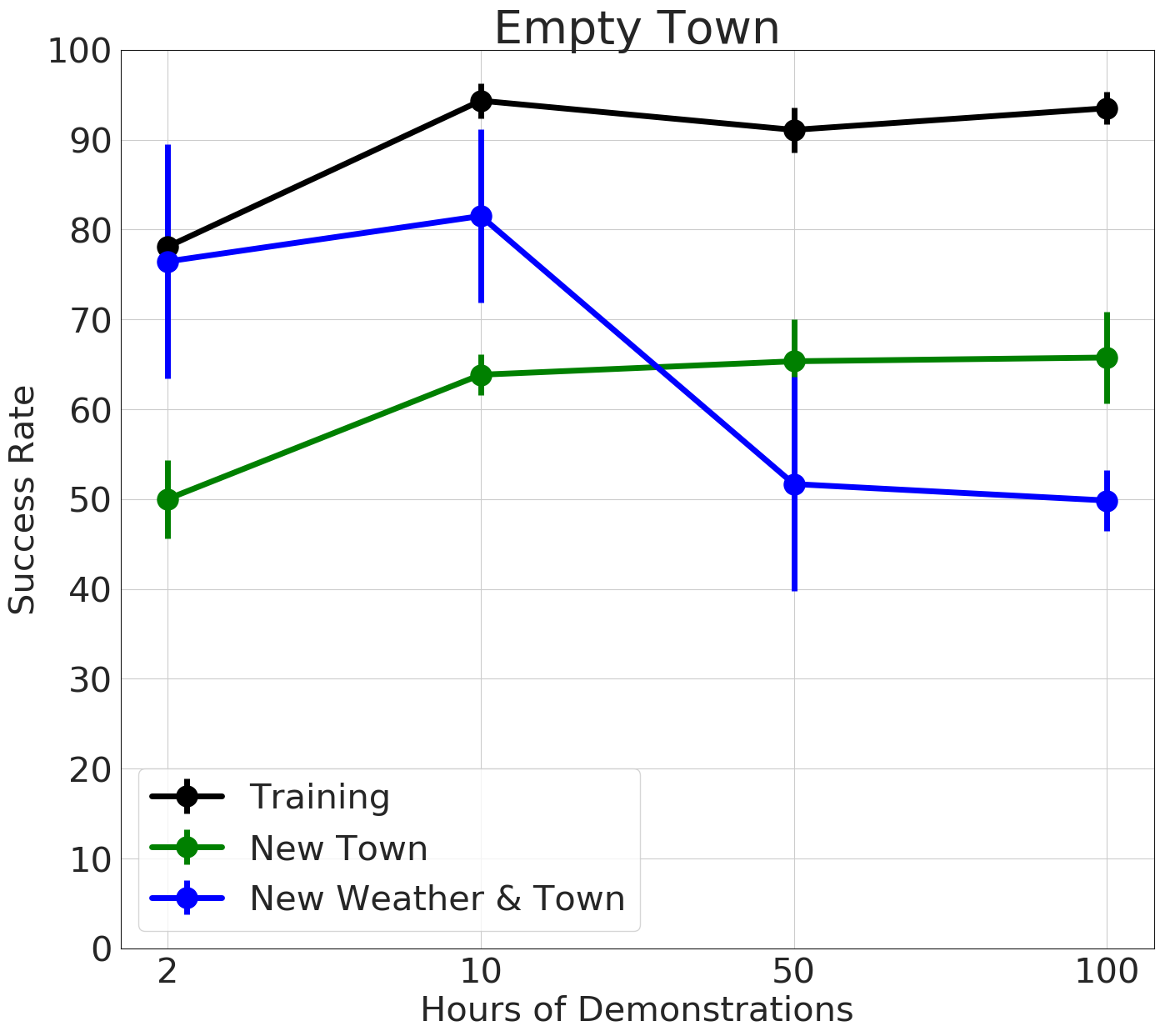} &
    \includegraphics[height=0.26\linewidth]{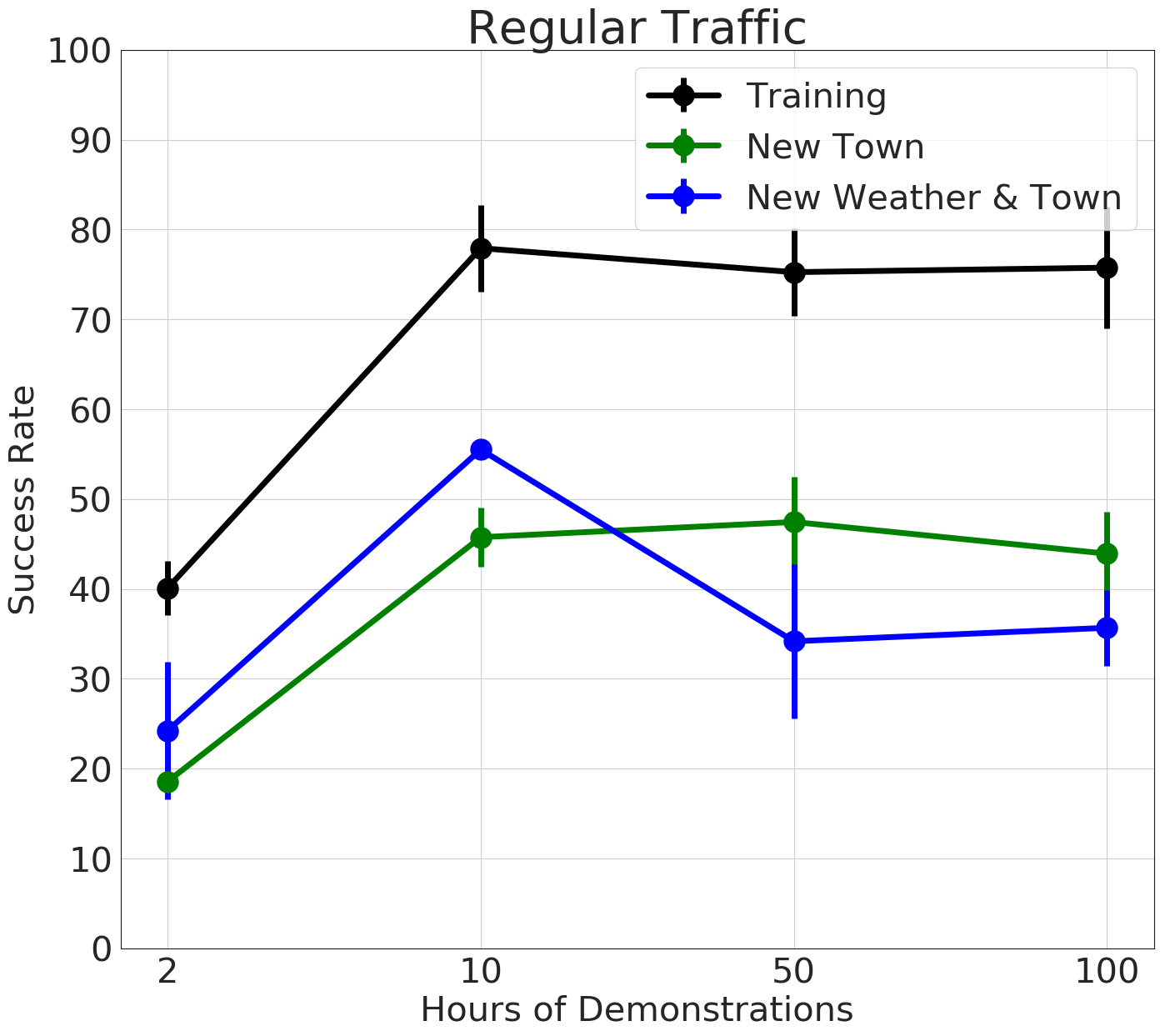} & \includegraphics[height=0.26\linewidth]{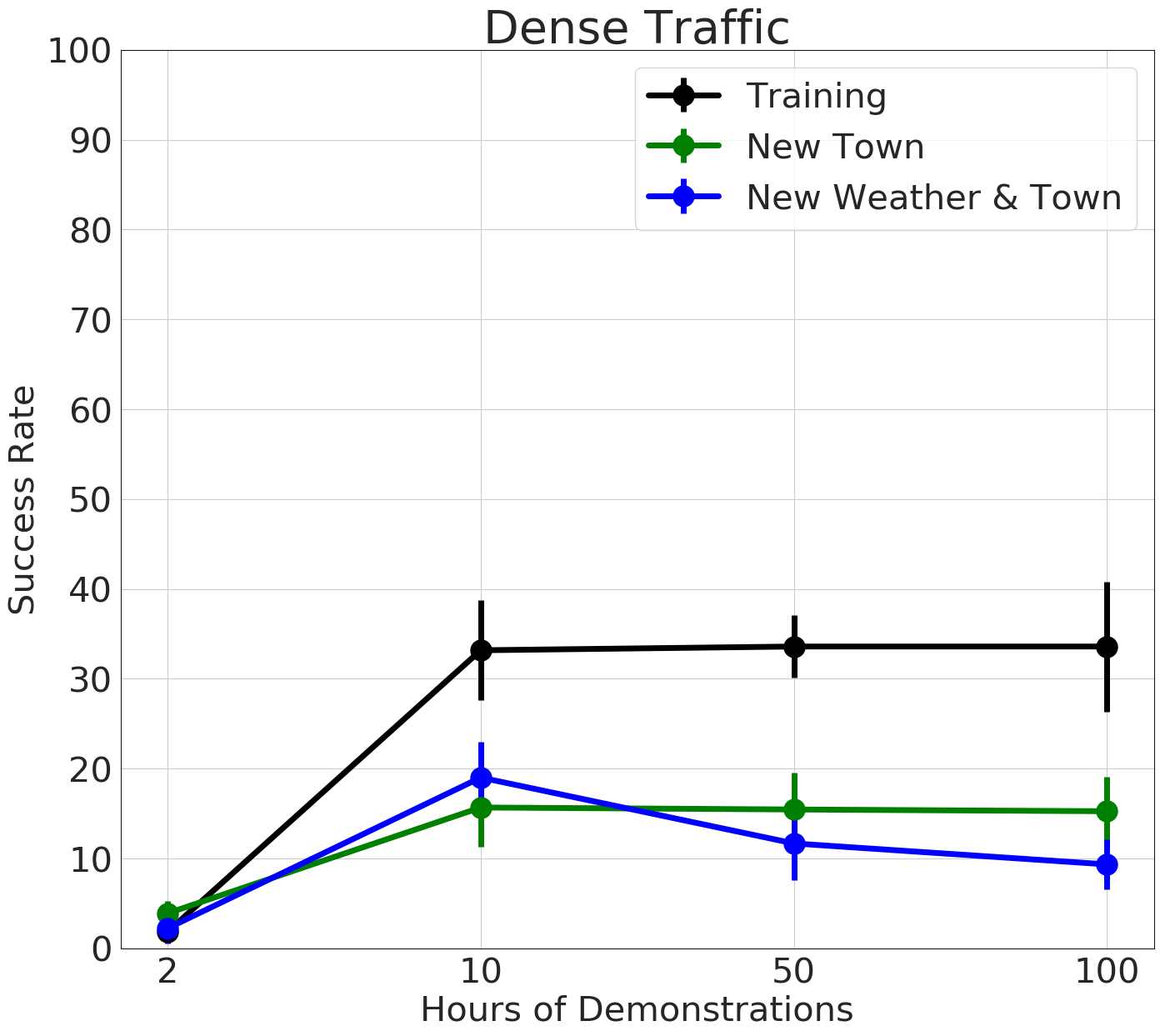}  \\ 

  \end{tabular}
  }
  \vspace{1mm}
  \caption{Due to biases in the data, the results may get either saturated or worse with increasing amounts of training data.}
  \label{fig:completion_data}
\end{figure*}

\begin{figure*}
 \centering
  
  {
  \setlength{\tabcolsep}{2.7pt}
  \begin{tabular}{ccc}
    \includegraphics[height=0.26\linewidth]{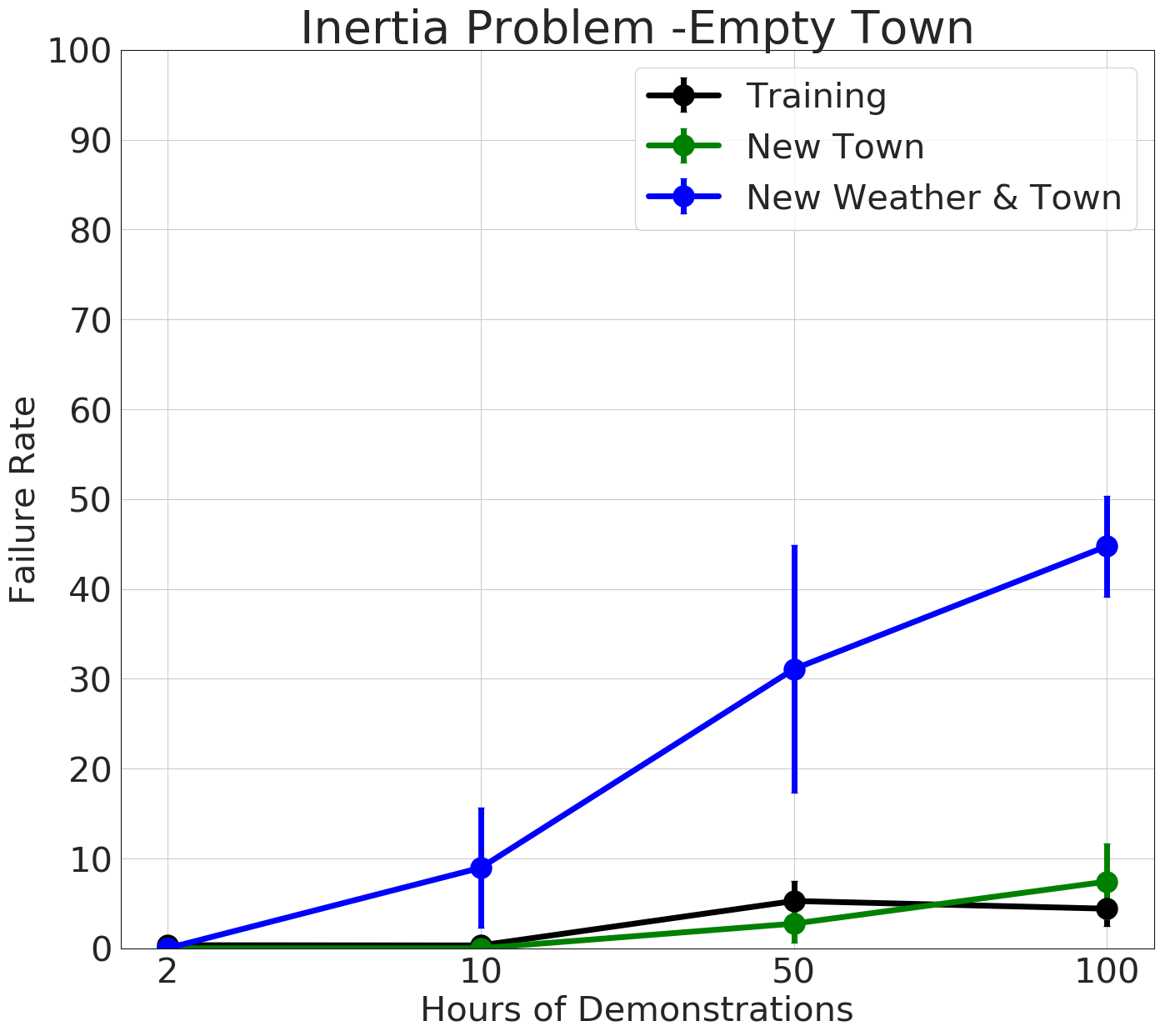} &
    \includegraphics[height=0.26\linewidth]{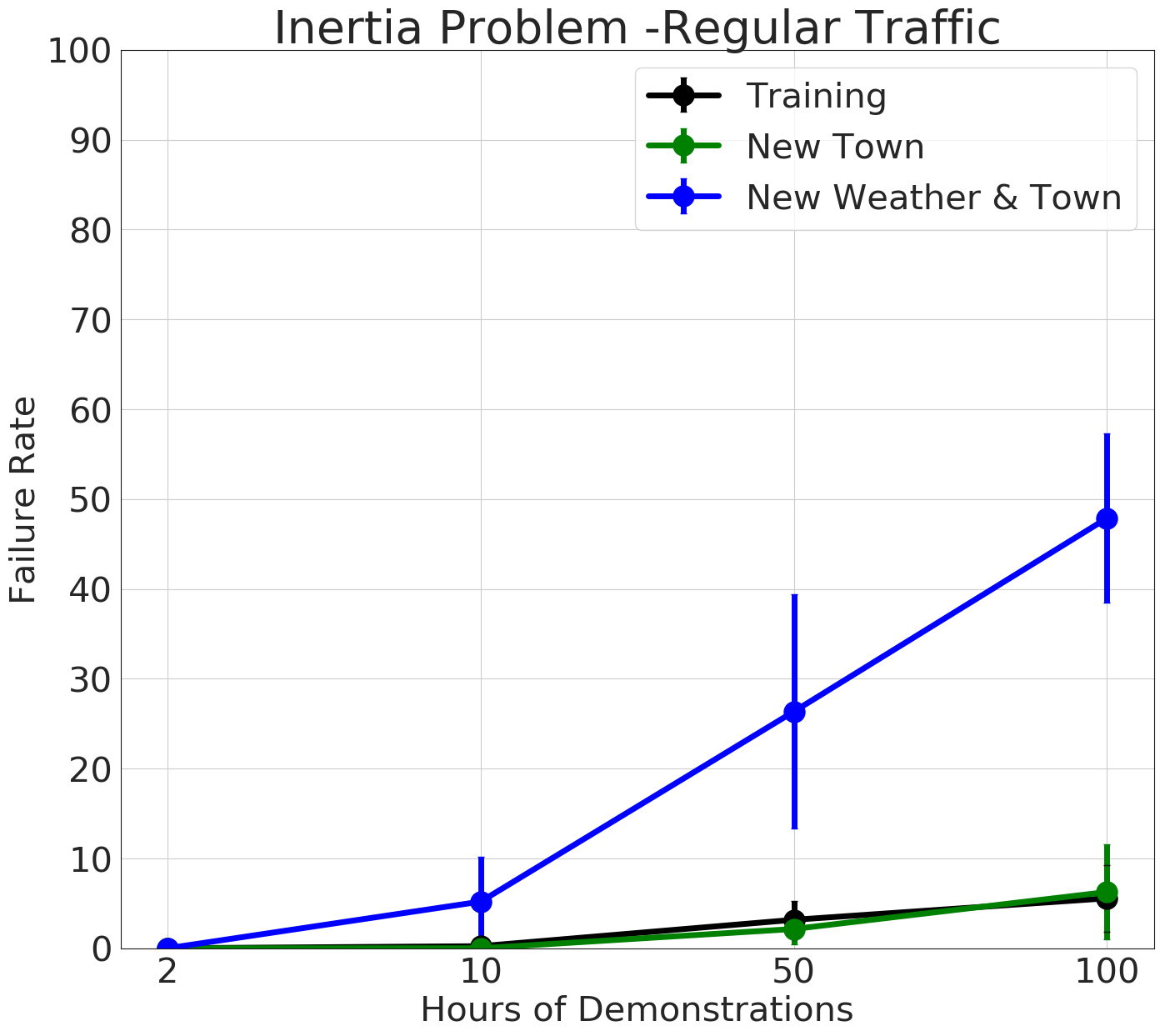} & \includegraphics[height=0.26\linewidth]{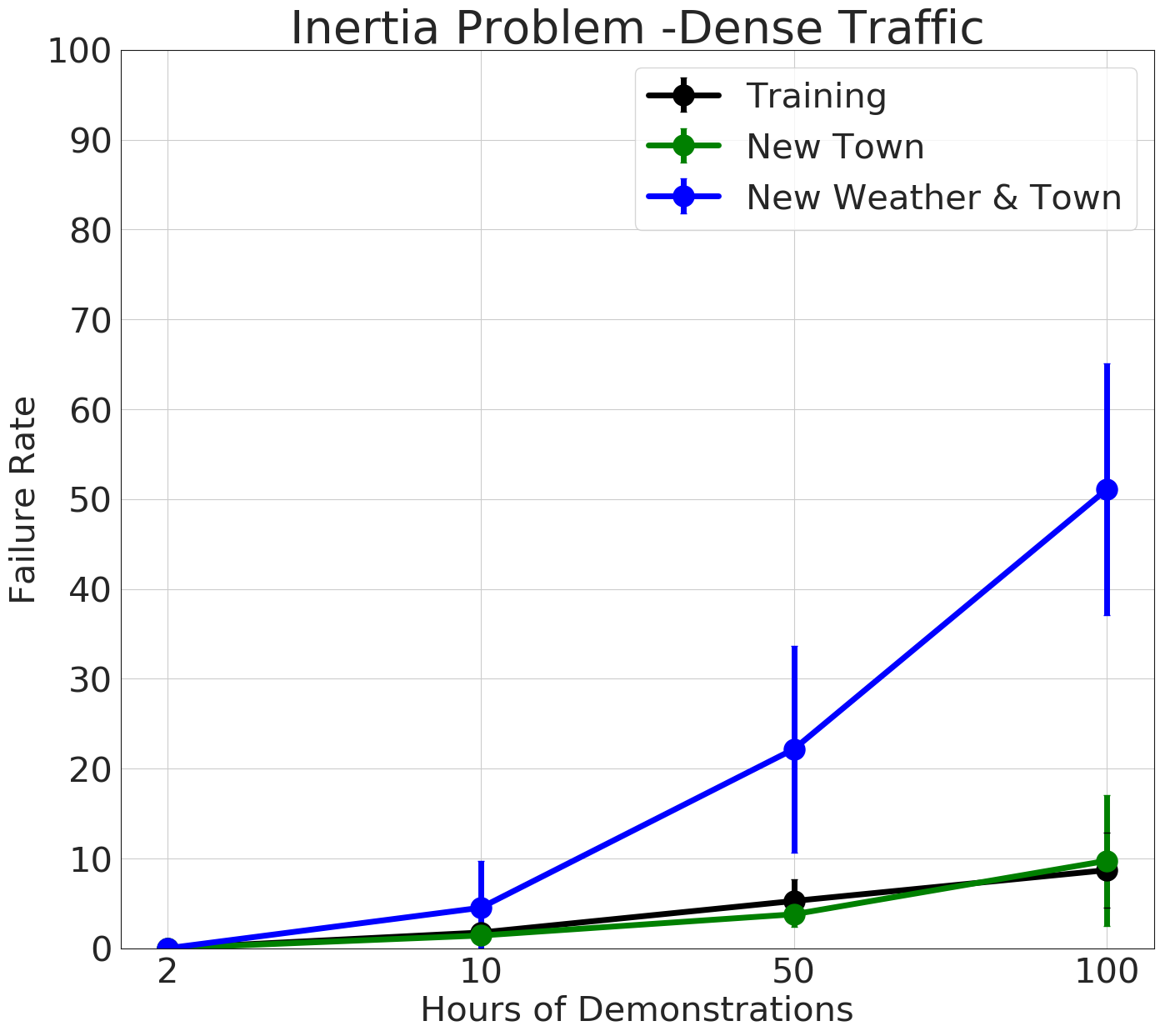}  \\ 

  \end{tabular}
  }
  \vspace{1mm}
  \caption{The percentage of episodes that failed due to the inertia problem. We can see that by increasing the amount of data, this bias may further degrade the generalization capabilities of the models.}
  \label{fig:completion_data_speed}
\end{figure*}

\begin{figure}[ht]
 \centering
  
  \includegraphics[height=0.70\linewidth]{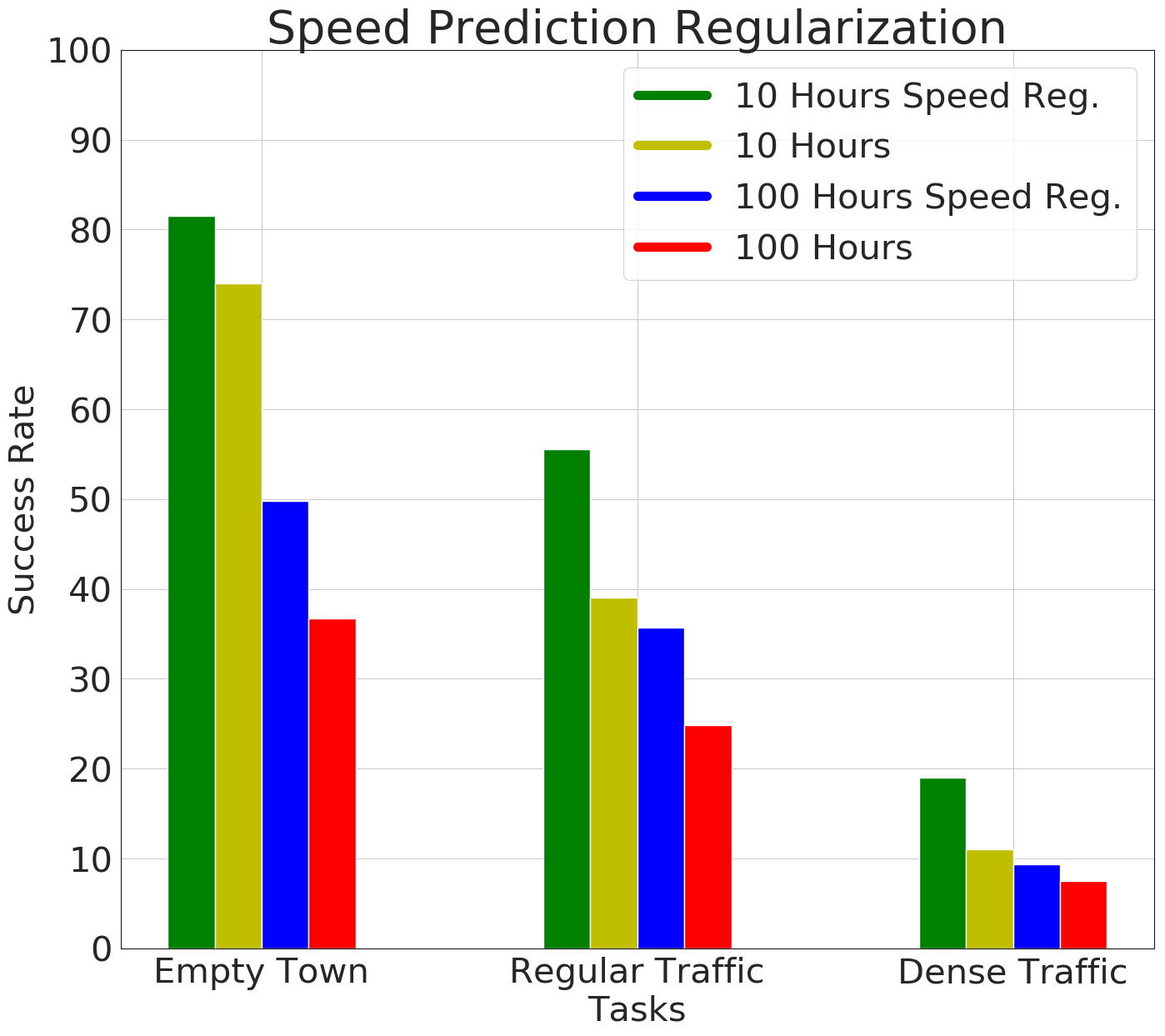} 
  \caption{ Comparison between the results with and without the speed prediction and different amounts of training
  demonstrations. We report
  the results only for the case were highest generalization is
  needed (New Weather and Town).}
  \label{fig:speed_pred_improvement}
\end{figure}

\vspace*{-2mm}
\paragraph{Driving Dataset Biases.} 

Figure~\ref{fig:completion_data} evaluates the effect of the amount of training demonstrations on the learned policy.
Here we compare models trained with 2, 10, 50 and 100 hours of demonstrations. The plots show the mean success rate and standard deviation over four different training cycles with different random seeds.
Our best results on most of the scenarios were obtained by using only 10 hours of training data, in particular on the ``Dense Traffic'' tasks and novel conditions such as New Weather and New Town.

These results quantify a limitation described in Section~\ref{sec:limitations_description}: the risk of overfitting to data that lacks diversity.
This is here exacerbated by the limited spatial extent and visual variety of our environment, including in terms of dynamic objects. We indeed observed that some types of vehicles tend to elicit better reactions from the policy than others. The more common the vehicle model and color, the better the trained agent reacts to it.
This raises ethical challenges in automated driving, requiring further research in fair machine learning for decision-making systems~\cite{barocas2017fairness}.
 



\vspace*{-2mm}
\paragraph{Causal confusion and the inertia problem.}

The main problem we observe caused by bias is the inertia problem
stemming from causal confusion, as detailed in Section~\ref{sec:limitations_description}. 
Figure~\ref{fig:completion_data_speed} shows the percentage of episodes that failed due to the agent staying still, without any intention to use the throttle, for at least 8 seconds before the timeout. Our results show the percentage of episodes failed due to that inertia problem increases with the amount of data used for training.
We proposed to use a speed prediction branch as part of our CILRS model (cf. Figure~\ref{fig:networkdiagram}) to mitigate this problem.
Figure~\ref{fig:speed_pred_improvement} shows the percentage of successes for the New Weather \& Town conditions on different tasks with and without speed prediction.
We observe that the speed prediction branch can substantially improve the success rate thanks to its regularization effect. It is, however, not a final solution to this problem, as we still observe instances
of the inertia problem after using this approach.


\begin{figure*}
 \centering
  
  {
  \setlength{\tabcolsep}{2.7pt}
  \begin{tabular}{cc}
    \includegraphics[height=0.42\linewidth]{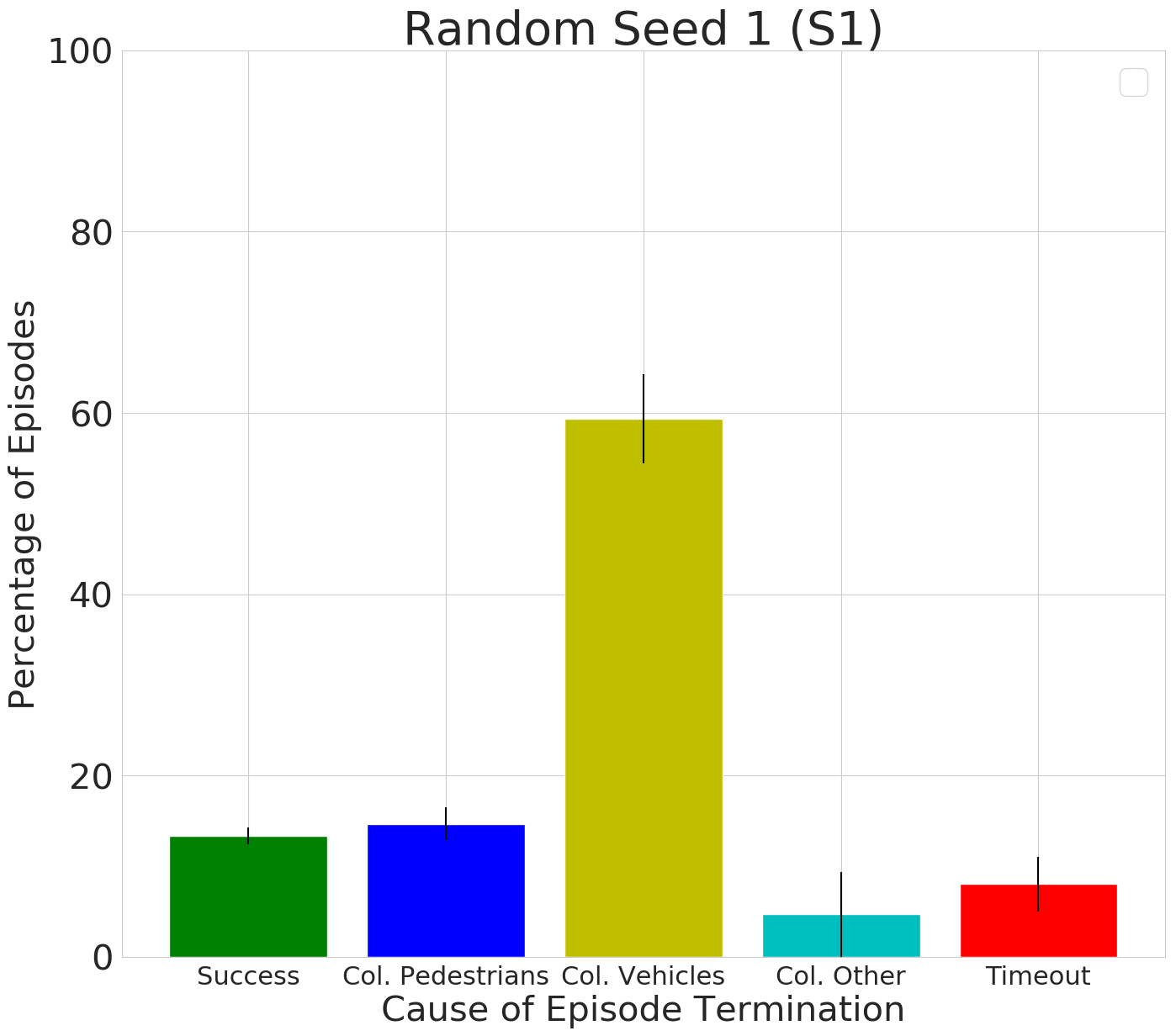} &
    \includegraphics[height=0.42\linewidth]{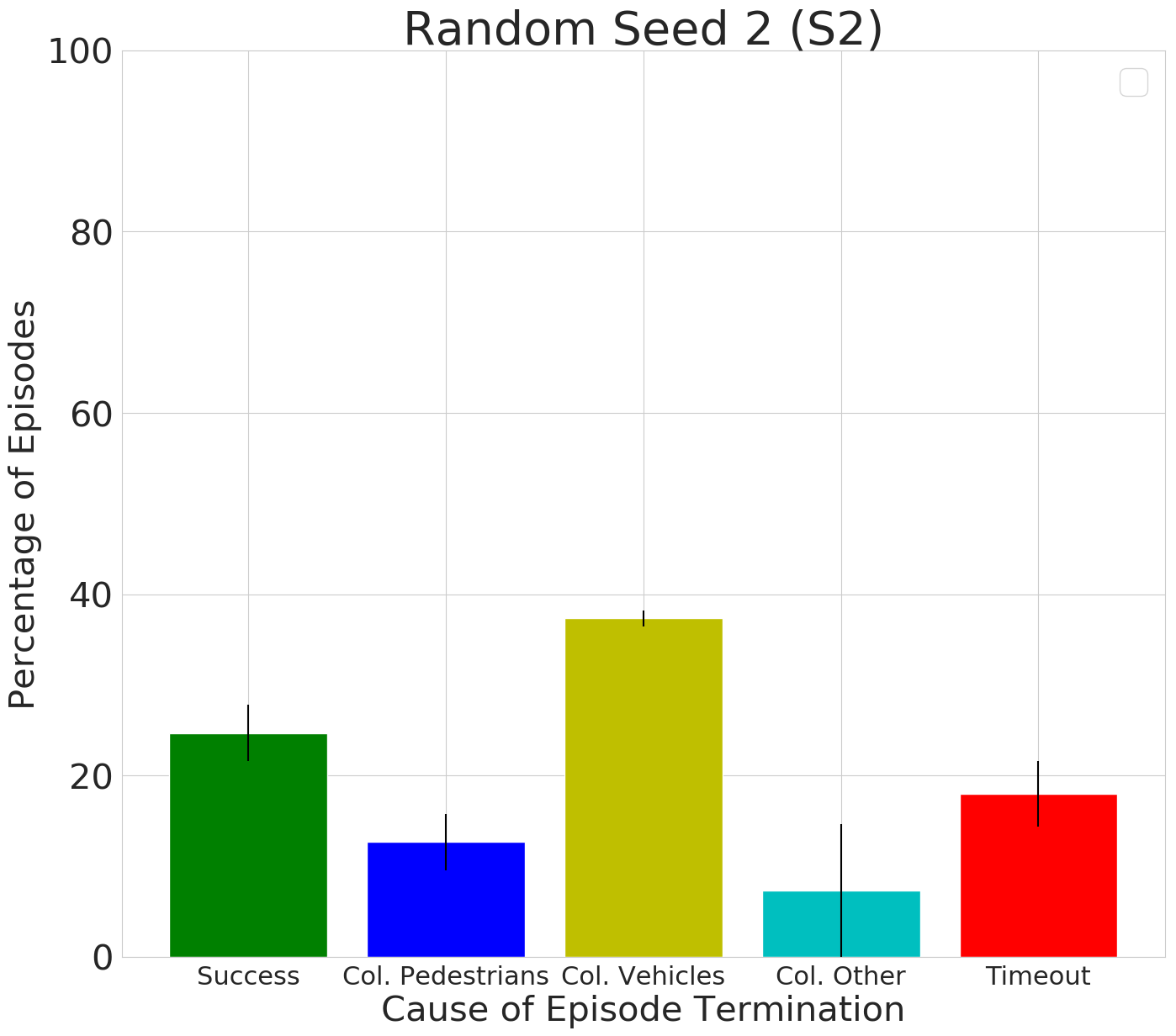}

  \end{tabular}
  }
  \vspace{2mm}
  \caption{Cause of episode termination on \emph{NoCrash} for two CILRS models (trained on 10 hours with ImageNet initialization) with identical parameters but different random seeds. The episodes were ran under ``New Weather \& Town'' conditions of the ``Dense Traffic'' task.
  }
  \label{fig:fine_comparison}
  \vspace{3mm}
  
\end{figure*}

\vspace*{-2mm}
\paragraph{High Variance.}

\begin{figure}[ht]
 \centering
  
  \includegraphics[height=0.50\linewidth]{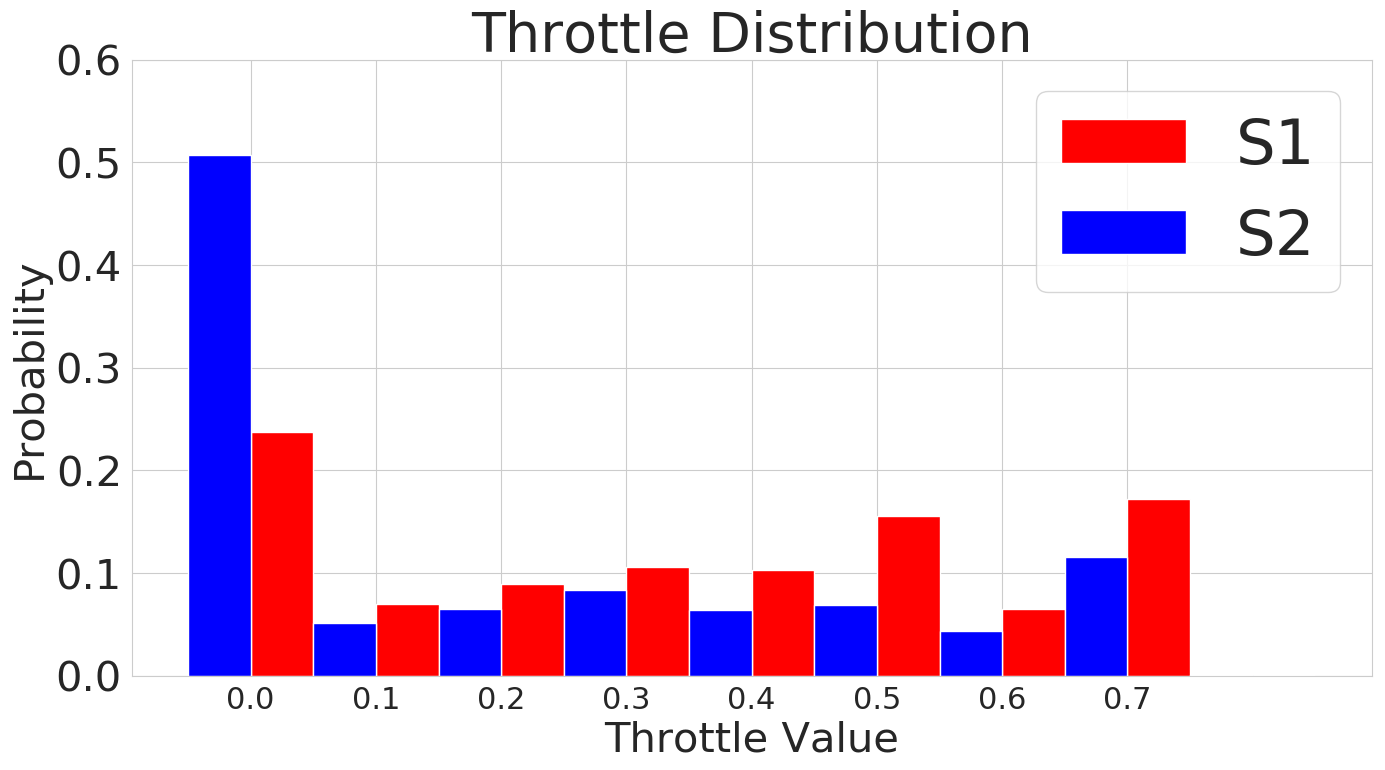}
  \caption{Probability distribution of having certain throttle values comparing models with two different random seeds but trained with the same hyper-parameters and data. We can see that S1 (red) is much more likely to have a higher throttle value.}
  \label{fig:histograms_throttle}
  \vspace{2mm}
\end{figure}

Repeatability of the training process is crucial for enhancing trust in end-to-end models. Unfortunately, we can still see drastic changes in the learned policy performance due to the variance caused by initialization and data sampling (cf. Section~\ref{sec:limitations_description}).
Figure~\ref{fig:fine_comparison} compares the cause of episode termination for two models where the only difference is the random seed during training.
The Model S1 has a much higher chance of ending episodes due to vehicle collisions. Qualitatively, it seemed to have learned a less general braking policy and was more prone to rear-end collisions with other vehicles.
On the other hand, Model S2 is able to complete more episodes and is less likely to fail due to vehicle crashes. However, we can see that it times out more, showing a tendency to stop a lot, even in non threatening situations.
This can be seen by analyzing the histograms of the throttle applied by both models during the benchmark, as shown in Figure~\ref{fig:histograms_throttle}. We can see a tendency for throttles of higher magnitude on Model S1.

As off-policy imitation learning uses a static dataset for training, this randomness comes from the order in which training data is sampled and the initialization of the random weights. This can possibly define which minima the models converges to. 
Table~\ref{tbl:variability} quantifies the effect of initialization 
on the success rate of driving tasks by computing the variance expressed in Equation~\ref{eq:variance}. The expected policy
score was computed by averaging twelve different training runs. We also consider the variance with and without ImageNet initialization.
We can see that the success rate can change by up to $42\%$ for tasks with dynamic objects. ImageNet initialization tends to reduce the training variability, mainly due to smaller randomness on initialization but also due to a more stable learned policy.




\begin{table}[!t]
\footnotesize
\centering
\resizebox{0.9\linewidth}{!}{
 \begin{tabular}{@{}lcccc@{}}
\toprule
                    
    && Task &&  Variance \\
    \midrule
    \multirow{3}{*}{CILRS}&& Empty && $23\%$ \\
    && Regular && $26\%$ \\
    && Dense &&  $42\%$ \\
    \midrule
    \multirow{3}{*}{CILRS (ImageNet)}&& Empty && $4\%$ \\
    && Regular &&  $12\%$ \\
    && Dense && $38\%$ \\

\bottomrule
 \end{tabular}
}
\vspace{3mm}
\caption{Estimated variance of the success rate of CILRS
on \emph{NoCrash} computed by training 12 times the
same model with different random seeds. The variance is reduced by
fixing part of the initial weights with ImageNet pre-training.}
\label{tbl:variability}
\end{table}

\section{Conclusion}
\label{sec:conclusion}
Our new driving dataset (CARLA100), benchmark (\emph{NoCrash}), and end-to-end sensorimotor architecture (CILRS) indicate that behavior cloning on large scale off-policy demonstration datasets can vastly improve over the state of the art in terms of generalization performance, including mediated perception approaches with additional supervision. This is thanks to using a deeper residual architecture with an additional speed prediction target and good regularization.

Nonetheless, our extensive experimental analysis has shown that some big challenges remain open.
First of all, the amount of dynamic objects in the scene directly hurts all policy learning methods, as multi-agent dynamics are not directly captured.
Second, the self-supervised nature of behavior cloning enables it to scale to large datasets of demonstrations, but with diminishing returns (or worse) due to driving-specific dataset biases that require explicit treatment, in particular biases that create causal confusion (e.g., the inertia problem).
Third, the large variance resulting from initialization and sampling order indicates that running multiple runs on the same off-policy data is key to identify the best possible policies. This is part of the broader deep learning challenges regarding non-convexity and initialization, curriculum learning and training stability.

We will release CARLA100, the code of our demonstrator AI and CILRS model, as well as our \emph{NoCrash} benchmark to stimulate future research on these topics.

\section{Acknowledgements}

Felipe Codevilla was supported in part by FI grant 2017FI-B1-00162. Antonio M. L\'opez and Felipe Codevilla acknowledges the financial support by the Spanish TIN2017-88709-R (MINECO/AEI/FEDER, UE). Antonio M. López also acknowledges the financial support by ICREA under the ICREA Academia Program. As CVC/UAB researchers, Antonio and Felipe also acknowledge the Generalitat de Catalunya CERCA Program and its ACCIO agency. We also thank the generous TRI support in AWS instances to run the additional experiments after Felipe's internship. Special thanks to Yi Xiao for all the help on
making the video.

{
\bibliographystyle{ieee}
\bibliography{end-to-end-driving}
}

\clearpage

\section*{Appendix}
\label{sec:appendix}
\appendix{

\begin{table*}[t]
\small
\centering
\resizebox{1.0\linewidth}{!}{
 \begin{tabular}{@{}lcccccccccccccc@{}}
\toprule

       && \multicolumn{6}{c}{New weather} && \multicolumn{6}{c}{New town} \\
    Task             && CIL \cite{Codevilla2018}  & CIRL \cite{Liang2018CIRL}  & CAL \cite{Sauer2018conditional} & MT \cite{li2018rethinking} & CILR & CILRS && CIL \cite{Codevilla2018}  & CIRL \cite{Liang2018CIRL}  & CAL \cite{Sauer2018conditional} & MT \cite{li2018rethinking} & CILR & CILRS   \\
    \midrule
    Straight       && $98$ & $\textbf{100}$ & $\textbf{100}$ & $\textbf{100}$ & $96$ & $96$ && $97$ & $\textbf{100}$ & $93$ & $\textbf{100}$ & $92$ & $96$  \\
    One Turn       && $90$ & $94$ & $\textbf{96}$  & $88$ & $\textbf{96}$ &$\textbf{96}$ && $59$ & $71$ & $82$  & $81$ & $81$ & $\textbf{84}$\\
    Navigation     && $84$ & $86$ & $90$  & $88$ & $94$ & $\textbf{96}$ && $40$ & $53$ & $70$  & $\textbf{72}$ & $60$ & $69$\\
    Nav. Dynamic   && $82$ & $80$ & $82$  & $80$ & $92$ & $\textbf{96}$ && $38$ & $41$ & $64$  & $53$ & $55$ & $\textbf{66}$\\

\bottomrule
 \end{tabular}
}
\vspace{1mm}
\caption{Comparison with the state of the art on the original CARLA benchmark for the conditions ``New Town'' and ``New Weather''. The table reports the percentage of successfully completed episodes in each condition.}
\label{tbl:comparison_legacy_new}
\end{table*}

\begin{table*}[t]
\small
\centering
\resizebox{1.0\linewidth}{!}{
 \begin{tabular}{@{}lcccccccccccccc@{}}
\toprule

       && \multicolumn{5}{c}{New Weather} && \multicolumn{5}{c}{New Town} \\
    Task             && CIL\cite{Codevilla2018}   & CAL\cite{Sauer2018conditional} & MT\cite{li2018rethinking} & CILR & CILRS  && CIL\cite{Codevilla2018}  & CAL\cite{Sauer2018conditional} & MT\cite{li2018rethinking} & CILR & CILRS   \\
    \midrule

Empty && $83 \pm 2$ &$85 \pm 2$ &$58 \pm 2$ &$\textbf{98} \pm 1$ &$96 \pm 1$ && $48 \pm 3$ &$36 \pm 6$ &$41 \pm 3$ & $60 \pm 2$ & $\textbf{66} \pm 2$\\
Normal && $55 \pm 5$ &$68 \pm 5$ &$40 \pm 6$ & $69 \pm 4$ & $\textbf{77} \pm 1$  && $27 \pm 1$ &$26 \pm 2$ &$22 \pm 0$ & $42 \pm 2$ & $\textbf{49} \pm 5$ \\
Cluttered && $13 \pm 4$ &$33 \pm 2$ &$7 \pm 2$ & $27 \pm 3$ & $\textbf{47} \pm 5$ && $10 \pm 2$ &$9 \pm 1$ &$7 \pm 1$ & $12 \pm 2$ & $\textbf{23} \pm 1$ \\

\bottomrule
 \end{tabular}
}
\vspace{1mm}
\caption{ Comparison with the state-of-the-art on the \textit{NoCrash} Benchmark. Here we
compare on two extra conditions. New Weather refers to the same town as during training but with
new weather conditions. New Town refers to a town not seen during training.}
\label{tbl:comparison_nocrash_different}
\end{table*}

\section{CARLA100}
\label{sec:CARLA100}

Here we describe  the content of the CARLA100 dataset. Note that for training our model we only used RGB sensor data, the ego-vehicle forward speed, the high level turn intentions
for the conditional imitation learning and the ego vehicle controls.

\subsection{Expert Demonstrator}
\label{sec:expert}

We collect the dataset, here referred as CARLA100, by executing an automated navigation expert in the simulated environment. The expert has access to privileged information about the simulation state, including the exact map of the environment and the exact positions of the ego-car, all other vehicles, and pedestrians.
 
The path driven by the expert is calculated using a standard planner. This
planner uses an A* algorithm to determine the path to reach a certain
goal. This path is then converted into waypoints used by
a PID controller to generate the throttle, brake, and steering for 
the expert demonstrator.
The expert drives steadily on the center of the lane, keeping a constant speed of 35 km/h when driving straight and reducing the speed when making turns to about 15 Km/h. 

In addition, the expert is programmed to react to visible pedestrians when required to prevent collisions. The expert reduces its speed proportionally to the collision distance when the pedestrian is over 5 meters away and less than 15 meters away, or breaking to full stop when the pedestrian is less than 5 meters away. 

The proposed demonstrator also reduces its speed to follow lead cars.
The expert stops when the leading vehicle is closer than 5 meters.  For our data collection process the expert never performs lane changes or overtakes. 

To improve diversity, realism, and increase the number of visited state-action pairs, we add noise to the ego car controls. This reduces the difference between offline training and online testing scenarios \cite{Laskey2017noise}. We input noise to the expert demonstrator
in a similar way as proposed by \cite{Codevilla2018}. The noise simulates a gradual drift away from the desired trajectory of the experts. However, for training, the drift is not used, but only the reactions performed by the expert to correct the path. The added noise signal 
is detailed on Section \ref{sec:noise}

\subsection{Content}
\label{sec:content}

The dataset collection is divided into goal directed episodes
where the expert goes from a start position into a goal
position while stopping to avoid collisions with dynamic obstacles. In
total, we collected 2373 episodes with different characteristics.
The entire dataset was collected on Town01.
Each episode has the following features:
\begin{itemize}
    \item Number of Pedestrians: the total number of spawned pedestrians around the town. This number is randomly sampled
    from the interval $[50,100]$.
    \item Number of Vehicles: the total number of spawned vehicle around the town. This number is randomly sampled
    from the interval $[30,70]$.
    \item Spawned seed for pedestrians and vehicles: the random seed used for
    the CARLA object spawning process.
    \item Weather: the
    weather used for the episode is sampled from the set: \textit{Clear Noon, Clear Noon After Rain, Heavy Rain Noon, Clear Sunset}.
  
\end{itemize}
Each episode last from 1 to 5 minutes partitioned in simulation steps of 100 ms.
For each step, we store data divided
into two different categories: sensor data is
stored as PNG images, and measurement data is stored as json files.

For the sensor data we have the different camera sensors used:
RGB camera, and depth camera, and semantic segmentation pseudo sensor.
For each sensor we record data in three positions: aligned with
the car center, rotated 30 degrees to the left and rotated 30 degrees to the right.

As measurements, we have data measured from the ego-vehicle, the world status, and from all the other non player agents.
The following data was collected from the ego-vehicle and the world status:

\begin{itemize}
    \item Step Number: the simulation step that starts at zero and is incremented by one for every 100ms in game time.
    \item Game Time-stamp: the time that has passed since the simulation has started.
    \item Position: the world position of the ego-vehicle. It is expressed as a three dimensional vector $[x,y,z]$ in meters.
    \item Orientation: the orientation of the vehicle with respect
    to the world expressed as Euler Angles (row, pitch and yaw).
    \item Acceleration: the acceleration vector of the ego-vehicle
    with respect to the world.
    \item Forward Speed: the scalar speed of the ego vehicle in the forward direction of movement.
    \item Intentions: a signal that is proportional to the effect that the dynamic objects in the scene are having in the ego car actions. We use three different intention signals: stopping for pedestrians, stopping for cars and stopping for traffic lights. For example, an intention of 1 for stopping for pedestrian means that the ego car totally stopped for a pedestrian that is less than 5 meters away. An intention of the same class of 0.5 means that the expert noticed a pedestrians and has reduced its speed to a certain extent. An intention of 0 means there are no pedestrians nearby in the field of view of the expert.

    \item High Level Commands: the high level indication stating what the ego-vehicle should do in the next intersection: go straight, turn left, turn right, or do not care. Each of these commands are encoded as a
    integer number. $2$ is do not care, $3$ for turn left, $4$ for turn right, $5$ for go straight.
    \item Waypoints: a set containing the next 10 future positions the vehicle should assume. This is calculated with the path planning algorithm.
    \item Steering Angle: the current steering angle of the vehicle's steering wheel.
    \item Throttle: the current pressure on the throttle pedal.
    \item Brake: the current pressure on the brake pedal.
    \item Hand Brake: if the hand brake is activated not.
    \item Steer Noise: the current steering angle of the vehicle considering the noise function.
    \item Throttle Noise: the current pressure in the throttle pedal considering the noise function.
    \item Brake Noise: the current pressure in the brake pedal considering the noise function. The noise function is described in Section \ref{sec:noise}
    
\end{itemize}

For  each of the non-player agents (pedestrians, vehicles, 
traffic light), the following information is provided:

\begin{itemize}
    \item Unique ID: an unique identifier of this agent.
    \item Type: if it is a pedestrian, a vehicle or a traffic light.
    \item Position: the world position of the agent. It is expressed as a three dimensional vector $[x,y,z]$  in meters.
    \item Orientation: the orientation of the agent with respect
    to the world. Expressed as Euler angles (row, pitch and yaw).
    \item Forward Speed: the scalar speed of the agent in the forward direction of movement.
    \item State: only for traffic lights. Contains the state of the traffic light: either red, yellow or green.

\end{itemize}

\subsection{Noise Distribution}
\label{sec:noise}

During training data collection, $20\%$ of the time we injected noise into expert's steering signal.
Namely, at random point in time we added a perturbation to the steering angle provided by the driver.
The perturbation is a triangular impulse: it increases linearly, reaches the maximum value and then linearly declines.
This simulates smooth drift from the desired trajectory, similar to what might happen with a poorly trained controller.
The triangular impulse is parametrized by its starting time $t_0$, duration $\tau \in \Re^+$, sign $\sigma \in \{-1,\, +1\}$ and intensity $\gamma \in \Re^+$:
\begin{equation}
  s_{perturb}(t) = \sigma \gamma \max \left(0, \left(1 - \left|\frac{2(t-t_0)}{\tau} - 1 \right|\right)\right).
\end{equation}
Every second of driving we started a perturbation with probability $p_{perturb}$.
We used $p_{perturb} = 0.1$ in our experiments.
The sign of each perturbation was sampled at random, the duration was sampled uniformly from $0.5$ to $2$ seconds, the intensity was fixed to $0.15$.

\subsection{\textit{NoCrash} Benchmark}
\label{app:benchmark}

\begin{figure*}
  \centering
  { \fontsize{8pt}{10pt}\selectfont
  \setlength{\tabcolsep}{4pt}
  \begin{tabular}{cc}

    \includegraphics[width=0.45\linewidth]{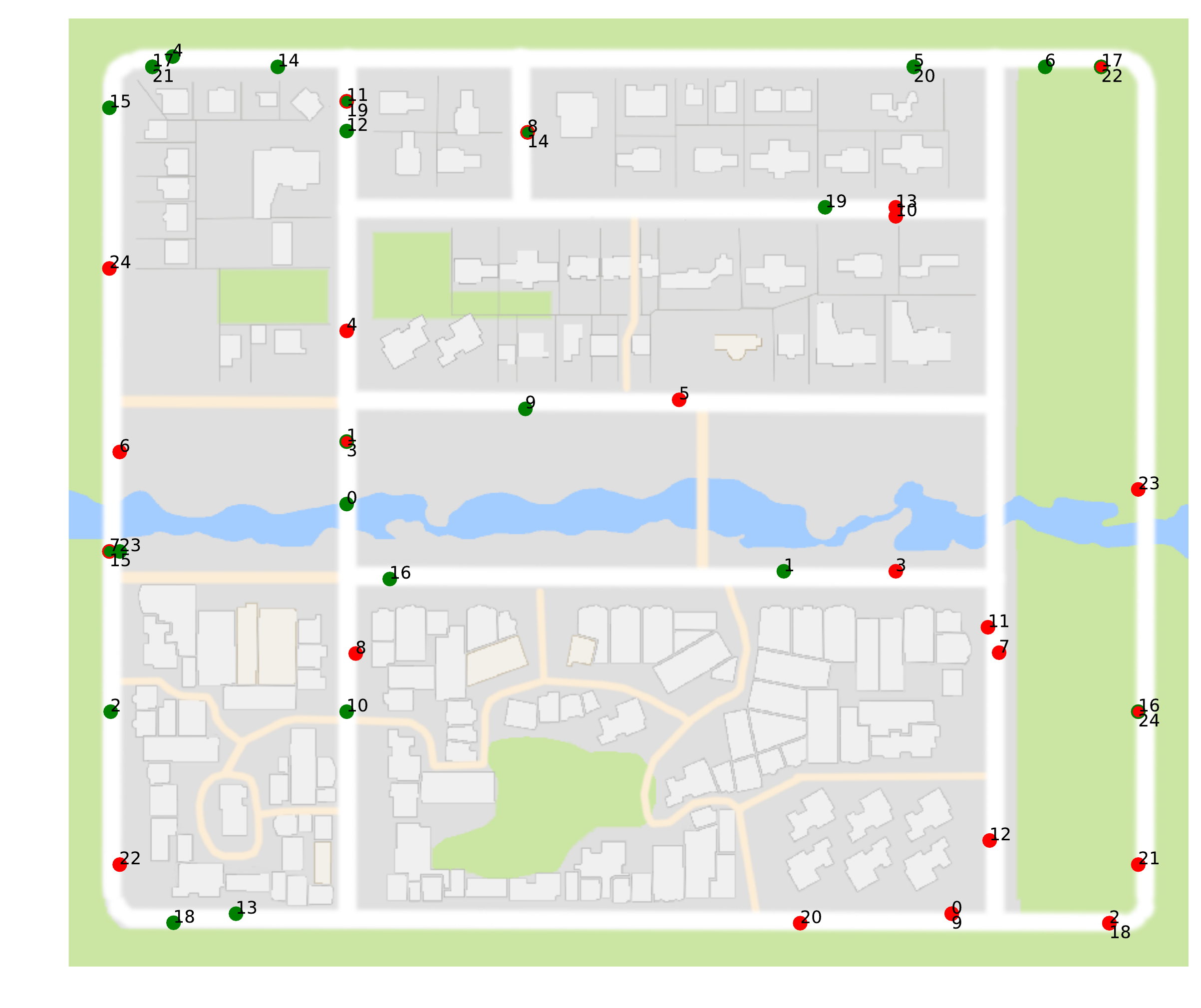} & \includegraphics[width=0.39\linewidth]{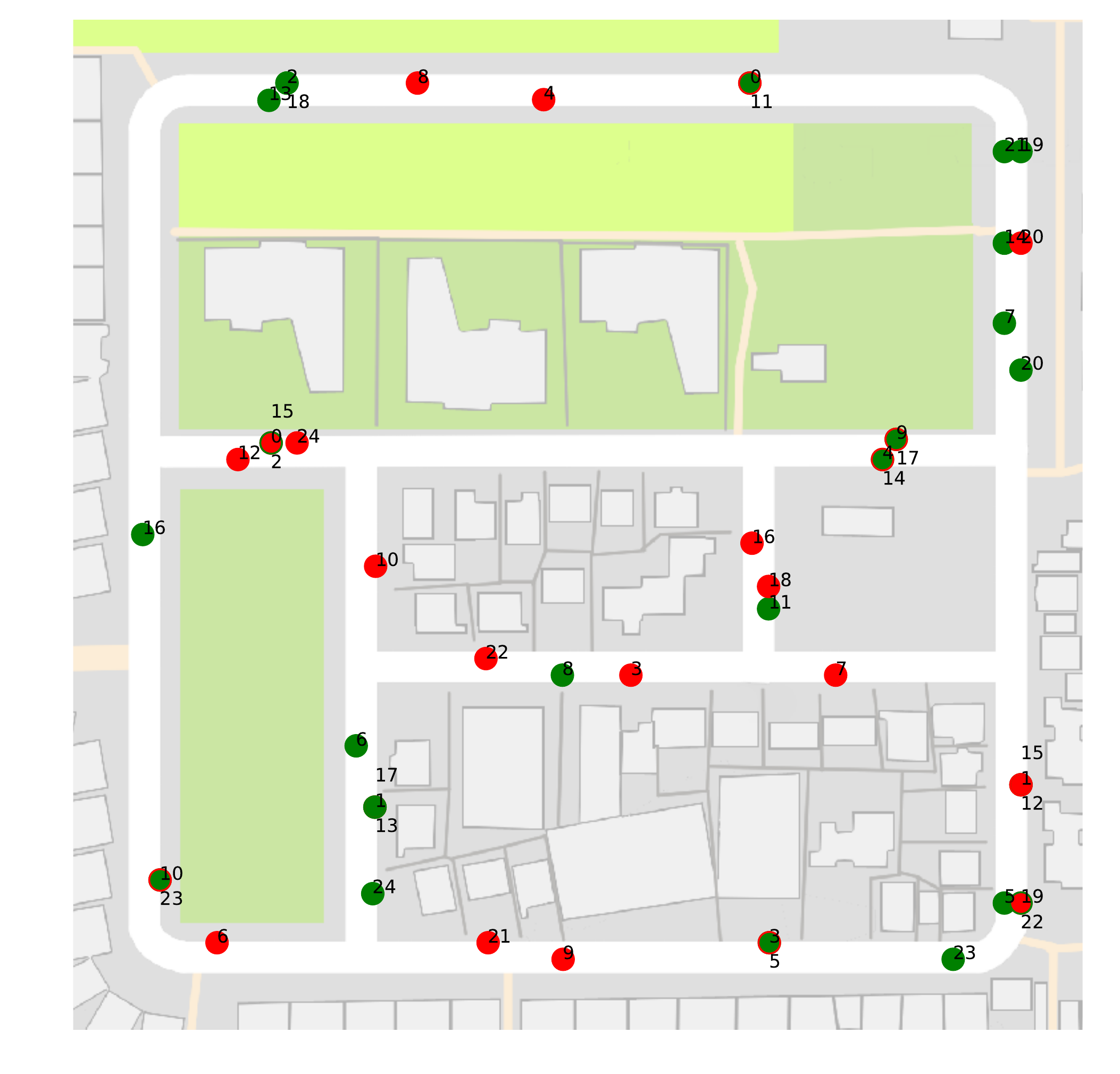} \\
    (a) Town01 &  (b) Town02

  \end{tabular}
  }
  \caption{The start  and goal positions for the CARLA100 Benchmark.
  The start positions are in red and the goal positions are in green. Same number correspond to matching start-goal positions.}
  \label{fig:positions}
  \vspace{-2mm}
\end{figure*}

The benchmark consist of three different tasks: ``Empty'', ``Regular'' and ``Cluttered''. The tasks
are better explained
on section 3 of the main text. Each task consists of 25 goal
directed episodes. In each episode, the agent is guided with a global planner
to reach a certain goal position. We consider an episode as a success if the agent reaches a certain goal under a time limit without colliding with any object, static or dynamic.
The tuples of start/goal positions are based on the ones used
in CARLA CoRL2017 benchmark for the tasks ``Navigation'' and
``Nav. Dynamic''. However, we removed some start-goal positions
that were too close to each other. Figure \ref{fig:positions} shows the start-goad positions for both Towns.

The benchmark is executed under four different conditions:

\begin{itemize}
    \item Training: The same one as collected on the training data.
    As mentioned above, we collected training data only in Town01 and using the weather conditions:
    ``Clear Noon'', ``Clear Noon After Rain'', ``Heavy Rain Noon'', ``Clear Sunset'
    \item New weather: The city as in the training data but with two different new weathers, ``After Rain Sunset'' and ``Soft Rain
    Sunset''.
    \item New Town: Same weathers as in ``Training'' but the tests take place
    in Town02.
    \item New Town \& Weather: Same weathers as in ``New Weather''
    but played in Town 02.
\end{itemize}

\section{Training Details}
\label{sec:apptraining}

\subsection{Architecture}

Table \ref{tbl:net_arch} details the standard architecture used in the experiments. For the perception module, we also experimented
with ResNet 18, ResNet 50 and  with the architecture proposed in \cite{Codevilla2018}.

		\begin{table}[]
		  \centering
			  \resizebox{0.8\linewidth}{!}{
		    \begin{tabular}{|c|cc|}
		      \hline
		      module                       & input dimension           & channels             \\ \midrule
		      Perception &   \multicolumn{2}{c|}{ResNet 34 \cite{he2016deep}  outputs 512}  		\\ \hline

		      \multirow{3}{*}{Measured Speed} & $1$                       & $128$             \\
		                                   & $128$                     & $128$                  \\
		                                   & $128$                     & $128$              \\  \hline
	                                   		      \multirow{3}{*}{Speed Prediction} & $512$                       & $256$             \\
	                                   & $256$                     & $256$                  \\
	                                   & $256$                     & $1$                  \\  \hline
		      \multirow{1}{*}{Joint input} & $512 + 128$               & $512$          \\ \hline
		      \multirow{3}{*}{Control}     & $512$                     & $256$      \\
											                 & $256$                     & $256$       \\
		                                   & $256$                     & $1$          \\ \hline
		    \end{tabular}
				}
		    \vspace{0.5cm}
		  \caption{Exact configurations of the architecture used on the experiments.}
		  \label{tbl:net_arch}
		\end{table}

\subsection{Image Input}
 Starting from a raw $800 \times 600$ pixels image, we cropped $125$ pixels from the top and $90$ at the bottom of the image and resized the resulting image to $200 \times 88$ pixels.

\begin{figure}[!ht]
 \centering
  
  {
  \setlength{\tabcolsep}{2.7pt}
  \begin{tabular}{ccc}
    \includegraphics[height=0.13\linewidth]{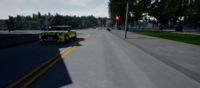} &
    \includegraphics[height=0.13\linewidth]{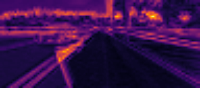} & \includegraphics[height=0.13\linewidth]{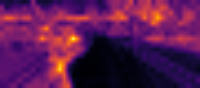}  \\ 
    \includegraphics[height=0.13\linewidth]{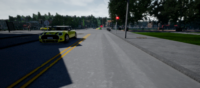} &
    \includegraphics[height=0.13\linewidth]{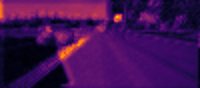} & \includegraphics[height=0.13\linewidth]{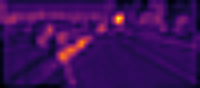}  \vspace{1mm} \\
    
    (a) Input Image & (b) Layer 1 & (c) Layer 2
    
  \end{tabular}
  }
  \caption{Activation maps showing the increased selectivity for traffic lights in the ResNet34 case (bottom) compared to the standard 8 convolution architecture (top). For the ResNet34, layer 1, refers to the attention maps  obtained after a full ResNet block.}
  \label{fig:traffic_lights}
  \vspace{-1mm}
\end{figure}



\section{Additional Results}
\label{app:additional}

\paragraph{Comparison}

Further comparisons with the state-of-the-art in the CARLA 
CoRL 2017 benchmark can be seen in Table  \ref{tbl:comparison_legacy_new}.
We can see that the ``New Town'' condition is  harder for some 
models than
the ``New Weather \& Town'' conditions. Yet the proposed methods can still outperform
previously proposed methods. One would expect ``New Weather \& Town'' to be the hardest task, but this discrepancy had been previously observed in the literature \cite{Dosovitskiy2017}.


\begin{figure*}
 \centering
  
  {
  \setlength{\tabcolsep}{2.7pt}
  \begin{tabular}{ccc}
    \includegraphics[height=0.26\linewidth]{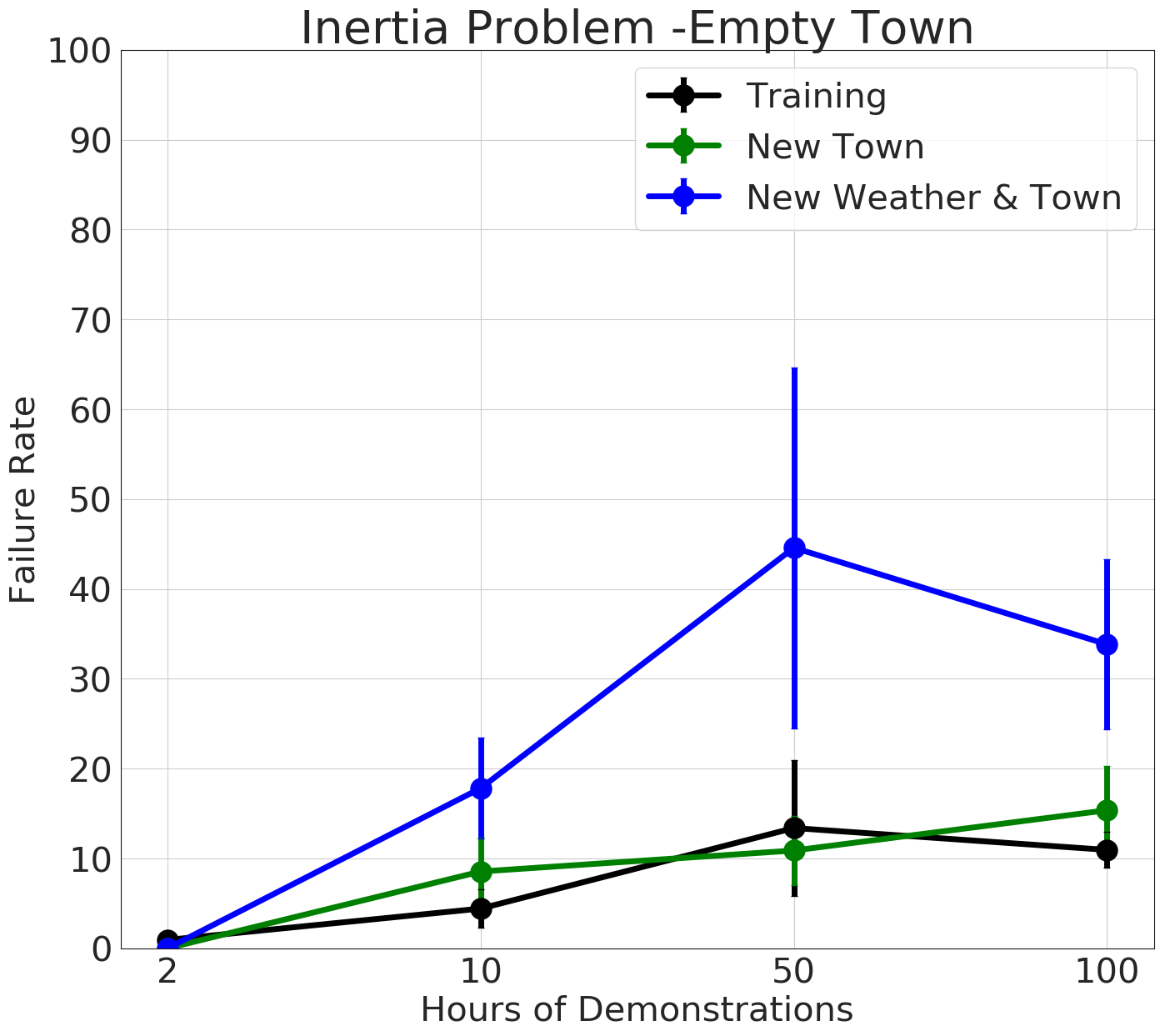} &
    \includegraphics[height=0.26\linewidth]{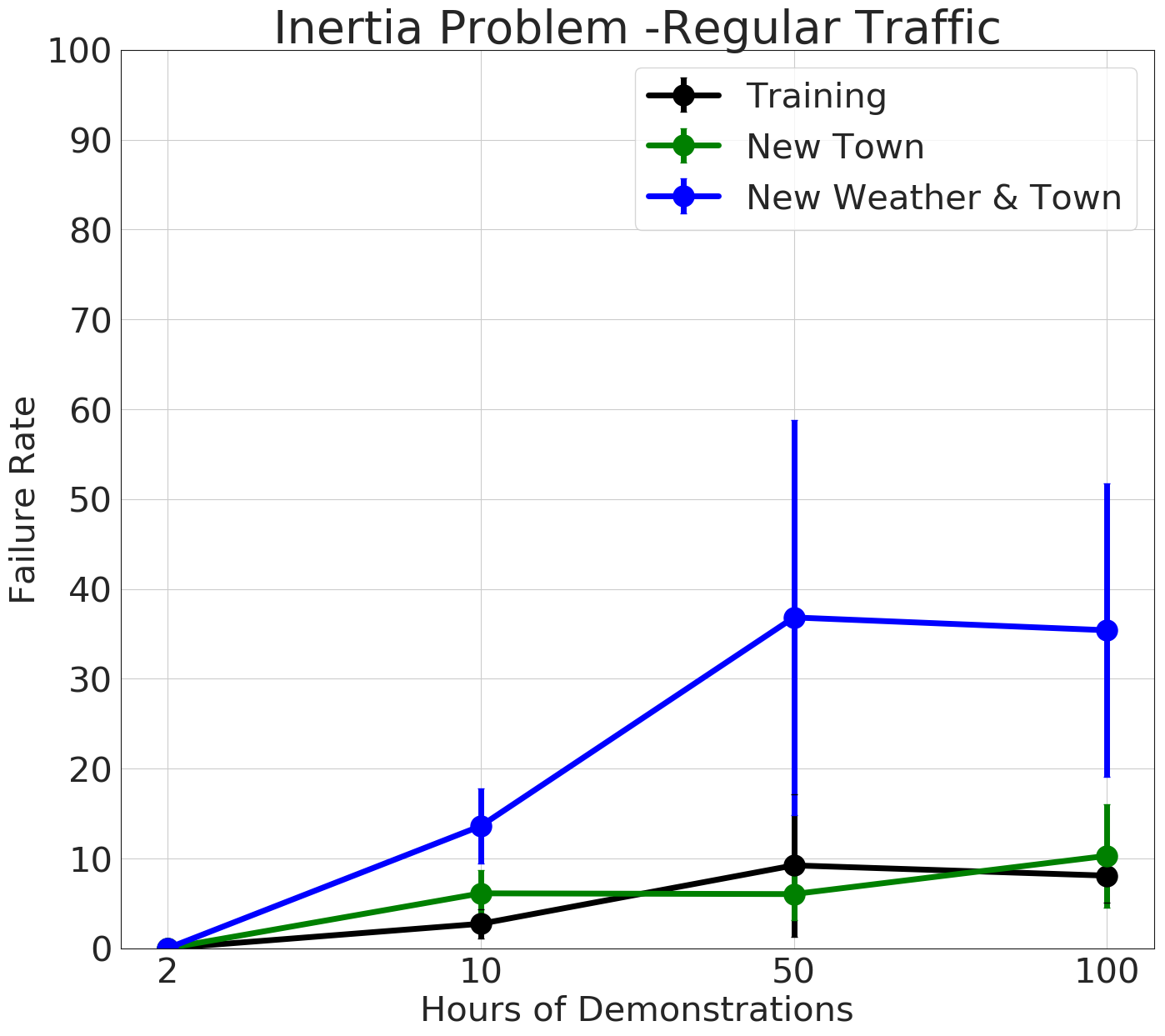} & \includegraphics[height=0.26\linewidth]{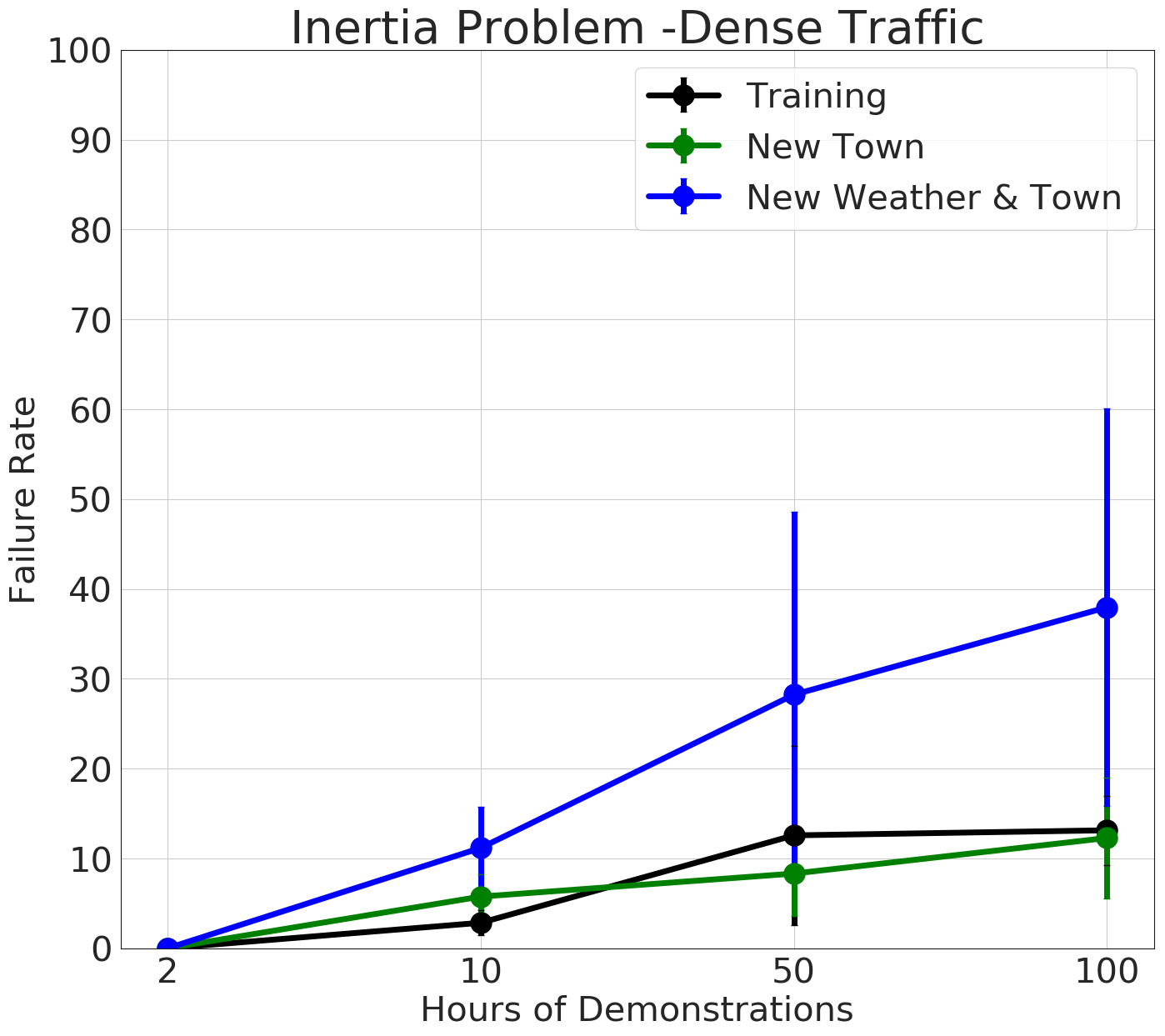}  \\ 

  \end{tabular}
  }
  \caption{Percentage of episodes ended by the ``inertia problem'' under different conditions. We report the mean and the standard deviation over four different training runs. We compare models with different amounts of training data and \textbf{without} image-net pre-training. We 
  can see that the inertia problem becomes more 
  prominent with more data.}
  \label{fig:completion_data_imnet_speed}
  \vspace{-1mm}
\end{figure*}

\begin{figure*}
 \centering
  
  {
  \setlength{\tabcolsep}{2.7pt}
  \begin{tabular}{ccc}
    \includegraphics[height=0.26\linewidth]{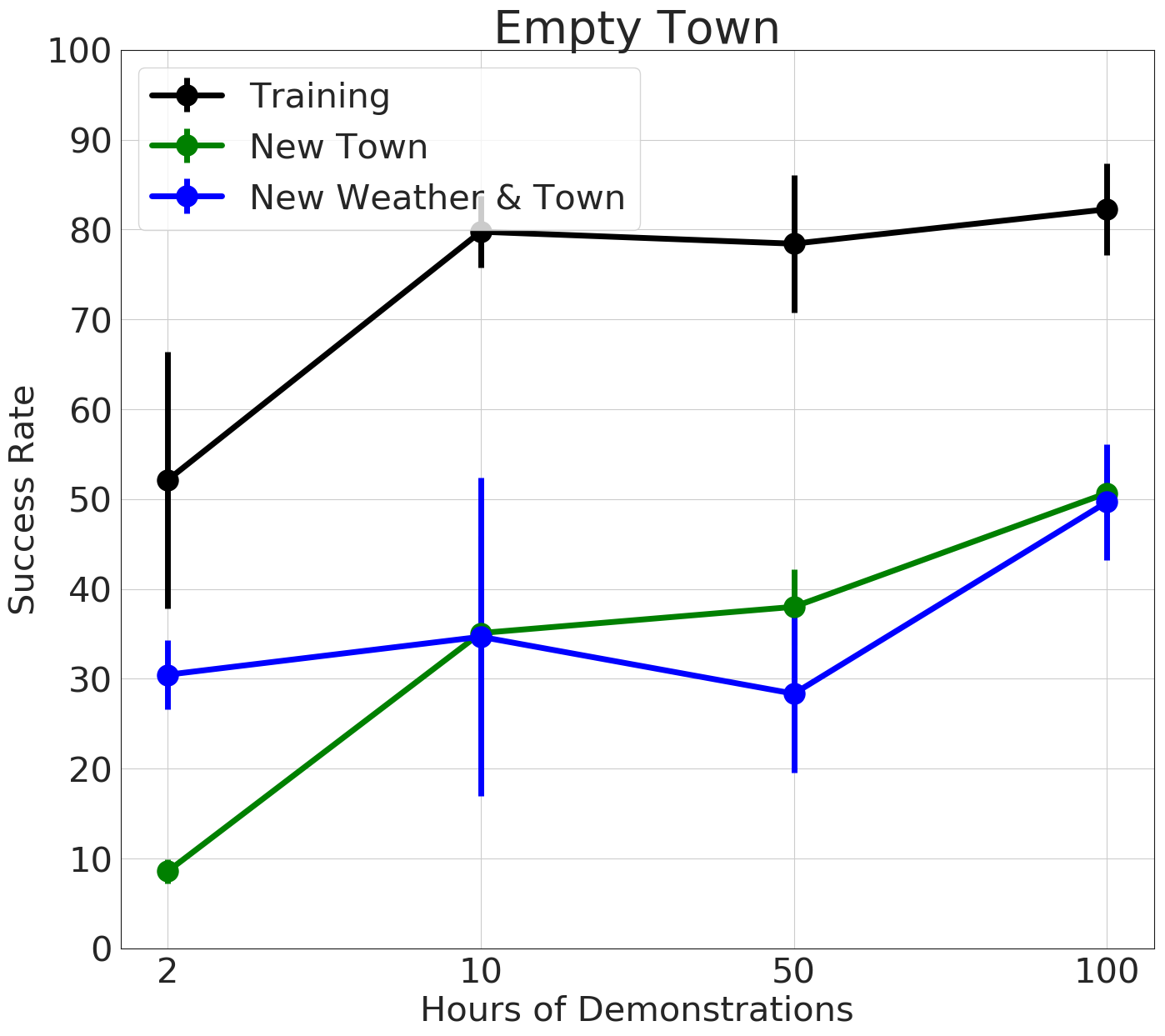} &
    \includegraphics[height=0.26\linewidth]{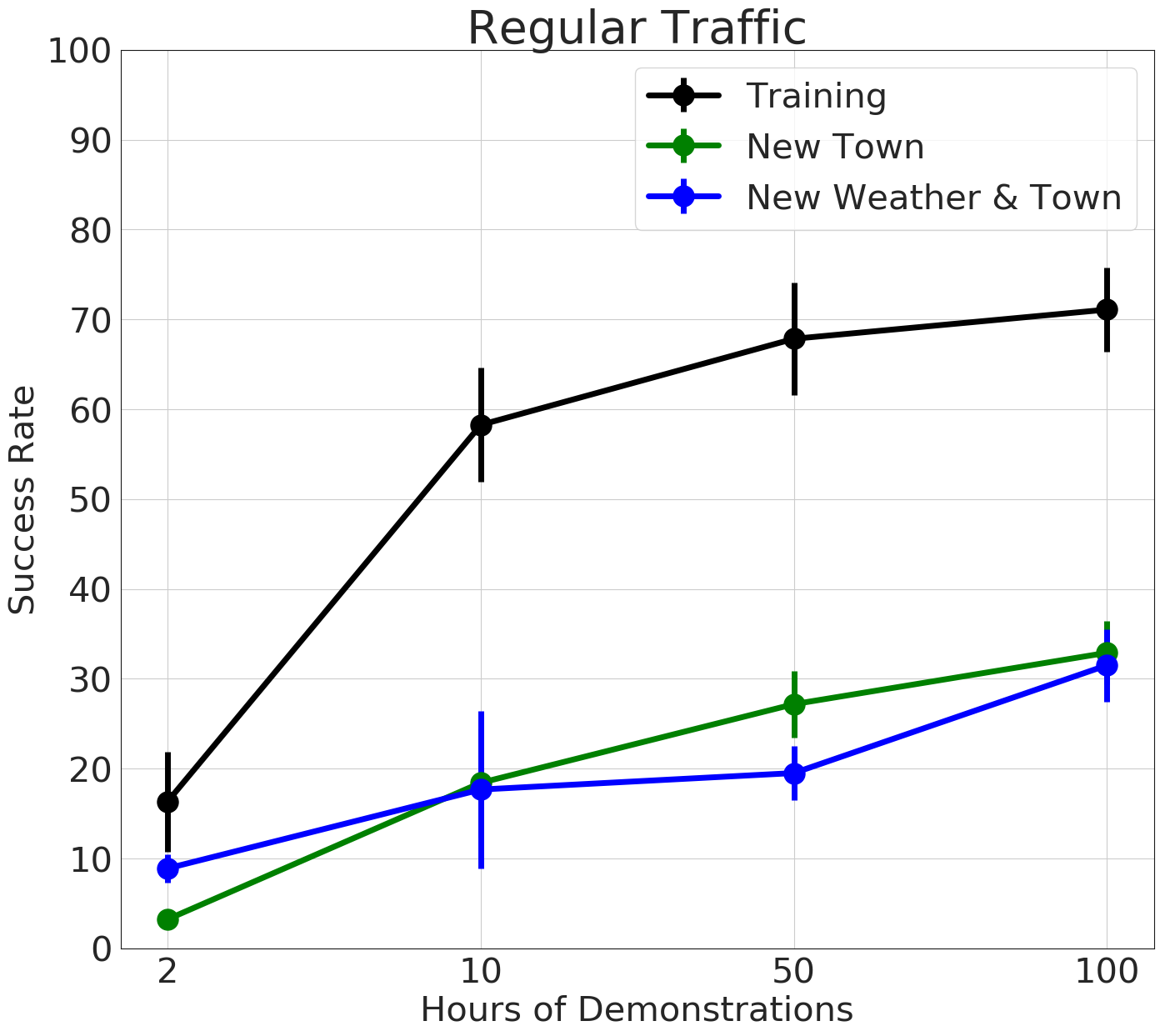} & \includegraphics[height=0.26\linewidth]{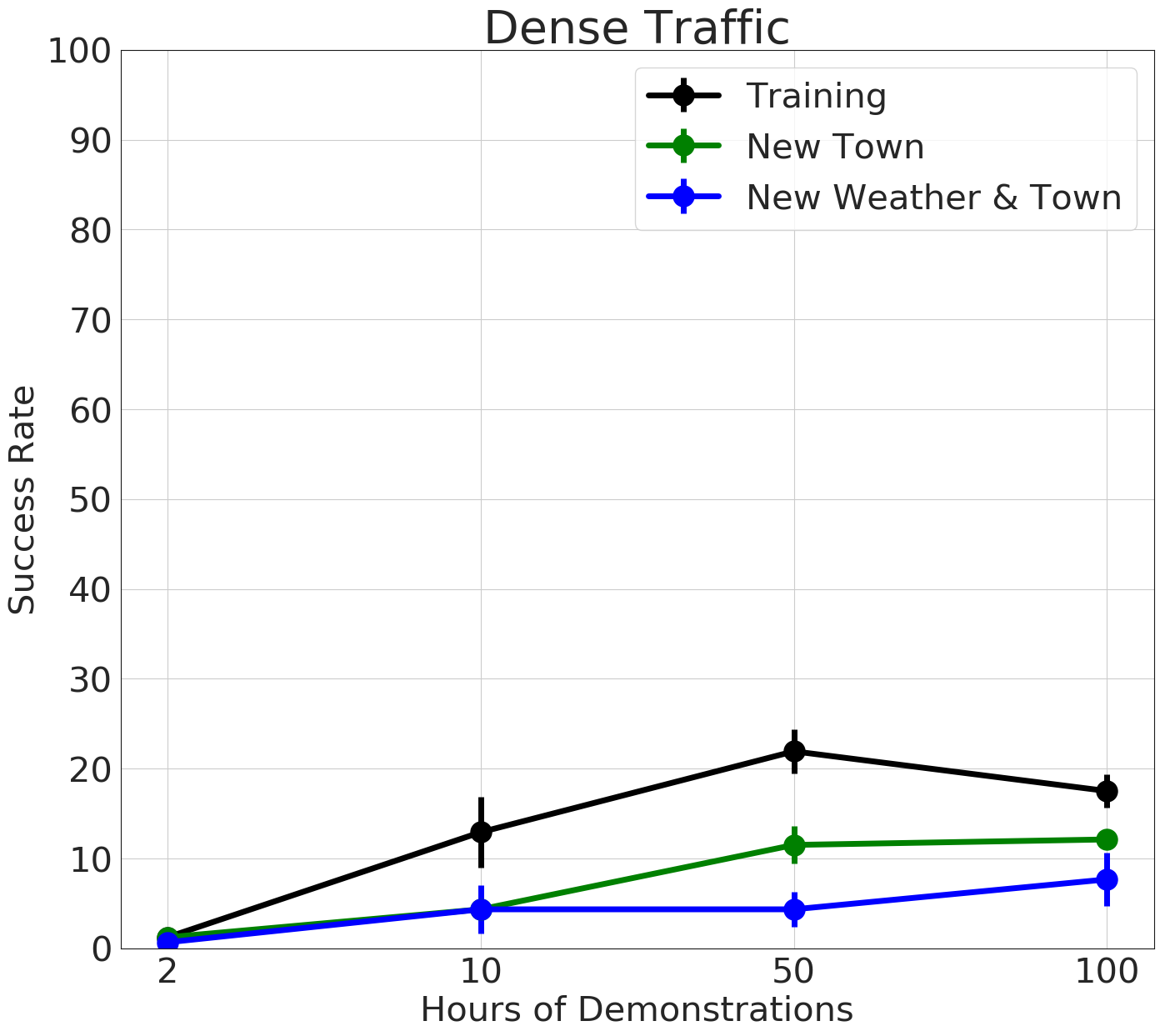}  \\ 

  \end{tabular}
  }
  \caption{The importance of data and initialization \textbf{without} ImageNet pre-training. We can see that the overall results improve with more data but not significantly. We
  can also see a case of worse performance when changing the ammount of training data from 50 to 100
  hours under the New Weather \& Town conditions with Dense Traffic.}
  \label{fig:completion_data_imnet}
  \vspace{-1mm}
\end{figure*}

\begin{figure*}
 \centering
  
  {
  \setlength{\tabcolsep}{2.7pt}
  \begin{tabular}{ccc}
    \includegraphics[height=0.26\linewidth]{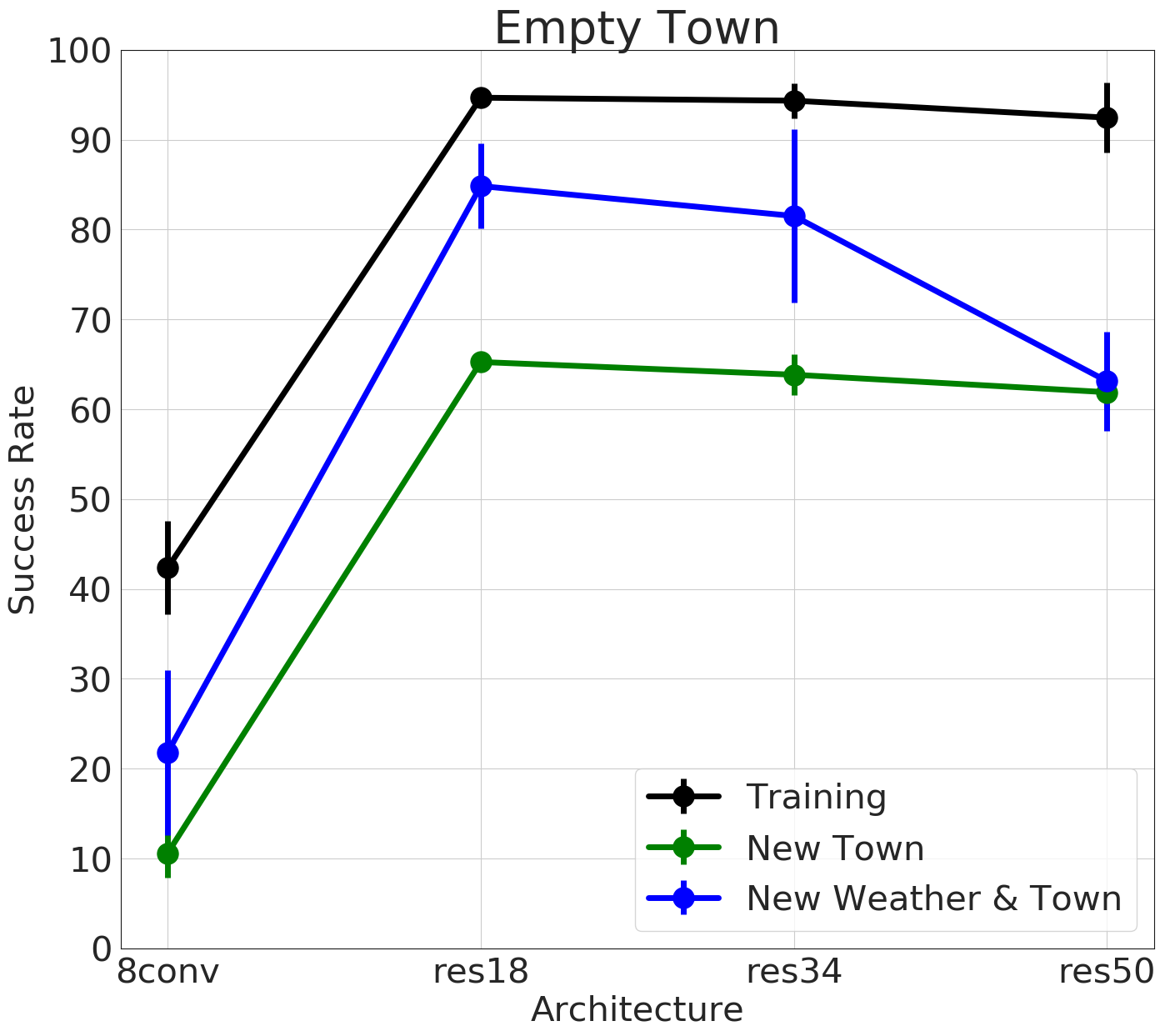} &
    \includegraphics[height=0.26\linewidth]{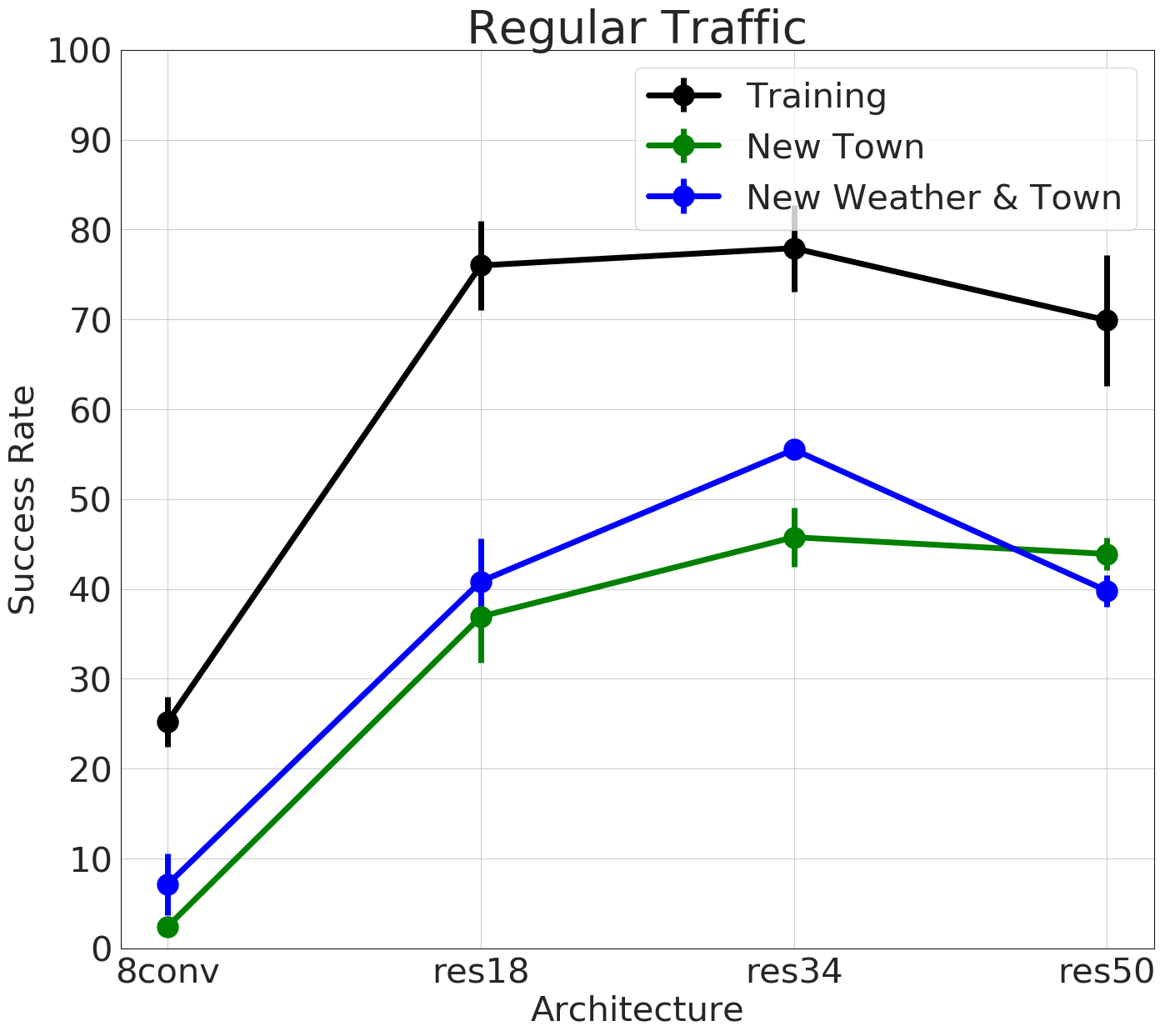} & \includegraphics[height=0.26\linewidth]{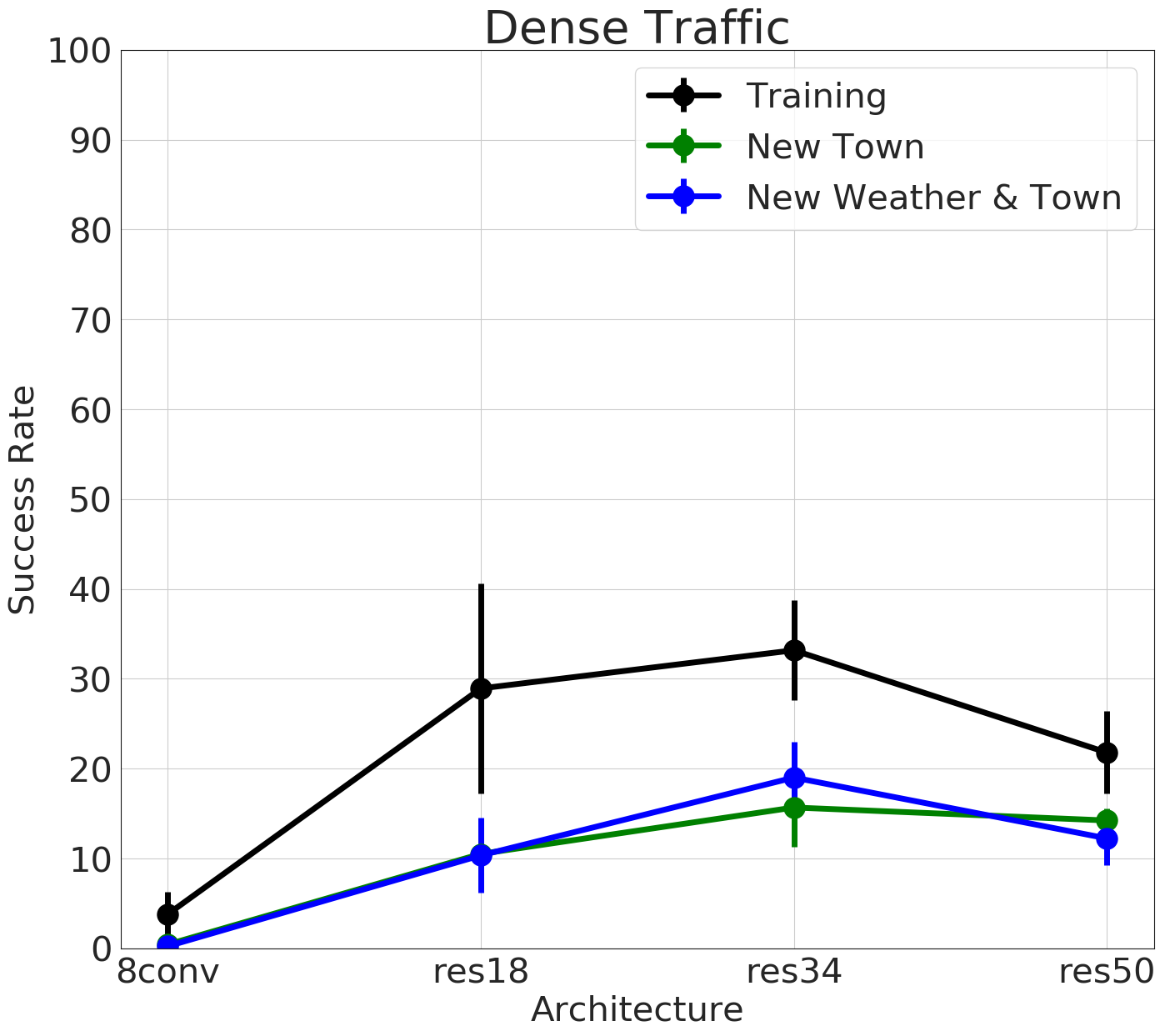}  \\ 

  \end{tabular}
  }
  \caption{Ablative analysis between different architectures. The eight convolutions architecture,``8conv'', proposed by Codevilla \cite{Codevilla2018}  obtained
  poor results on the more complex CARLA100 benchmark. ResNet based deeper architectures, ``res18'' and ``res34'',  were able
  to improve the results. However, when testing ResNet 50 we  notice a significant drop in the quality of the results.}
  \label{fig:completion_arch}
  \vspace{-1mm}
\end{figure*}

\paragraph{Changing Architecture}

In Figure \ref{fig:completion_arch}, we compare the results of the 8 layer convolutional model used by Codevilla et. al. \cite{Codevilla2018} and several new ResNet based configurations using our new dataset. 
First, we noticed that the 8 convolutions model obtained worse results than the ones reported as CIL at Table \ref{tbl:comparison_legacy}. 
This happened since we trained the 8 convolutions architecture with the more complex CARLA100 dataset. The model did not have enough capacity to capture the more complex actions and only fitted the several segments where the demonstrator stands still in front of traffic lights. This shows that higher capacity models 
  are able to better learn different sub-tasks. However, the results get worse for the deeper ResNet50 based models, showing that there is still possibility for different types of overfitting.

\paragraph{ImageNet Initialization}

We also observed a considerable change in the success rate results when not using ImageNet initialization, as show in Figure \ref{fig:completion_data_imnet}.
Without Imagnet initialization, the highest success rates obtained by models trained on 100 hours of demonstrations. However, these later results are still below what can be achieved  with less data and ImageNet pre-training, specially on the dense traffic tasks.

\subsection{Reacting to Traffic Lights}




We show that interesting policies around traffic lights emerged for some of the models we trained. 
In Table \ref{tbl:traffic_lights}
we show the percentage of traffic lights that were crossed on green light 
for different models. This number is computed for the ``Empty Town'' task from the 
dynamic urban scenarios benchmark. 
The original CIL model trained with older data \cite{Codevilla2018} represents an effective policy that was trained without demonstrations of stopping for red traffic light, so its number can be seen as a lower bound. The $8$ convolutions is a model with the same architecture as the CIL model but trained
to react to traffic lights with the proposed new dataset. We can see that this model was more reactive 
to traffic lights, but very poorly. On the other hand, our best model, having only 47\% of traffic light violations, is clearly stopping for a significant amount of red traffic lights.
This result is even more expressive considering the version using 100 hours of training data which did only 27\% of traffic light violations.  
However, when we analyze generalization conditions, Tab. \ref{tbl:traffic_lights} bottom, we see there is an ample room for improvement. Regardless, such improvement in this longitudinal controls task is promising for modeling lateral and longitudinal controls jointly end-to-end.

\begin{table}[!t]
\small
\centering
\resizebox{1.0\linewidth}{!}{
 \begin{tabular}{@{}lccc@{}}
\toprule
    Condition & Models  &&  Traffic Light Violations  \\
    \midrule
    \multirow{4}{*}{ Training Conditions }& CIL     && $83$\\
    & 8 Conv  && $71$ \\
   & CILRS 10 && $47$ \\
   & CILRS 100 && $\textbf{27}$ \\
     \midrule
   \multirow{4}{*}{New town \& weather} & CIL  && $82$ \\
   & 8 Conv && $81$ \\
   & CILRS 10 hours && $\textbf{64}$ \\
   & CILRS 100 hours && $78$ \\
\bottomrule
 \end{tabular}
}
\vspace{1mm}
\caption{Percentage of times the agents crossed a traffic light on red (lower is better) in the ``Empty'' conditions of the \textit{NoCrash} benchmark.}
\label{tbl:traffic_lights}
\end{table}

\subsection{Main Causes of Failure}

On Table \ref{tbl:cause_of_failure} we show some of our models compared with some of
the literature with regard to their cause of failure. We
specify the percentage of episodes that ended due to different causes 
of crash, due to timeout of the task or if the main cause is that the controller
stopped and never resumed moving again (i.e the inertia problem).

\begin{table*}[!t]
\small
\centering
\resizebox{1.0\linewidth}{!}{
 \begin{tabular}{@{}lccccccccclcccccc@{}}
\toprule

                    &&  && \multicolumn{5}{c}{Training} && \multicolumn{5}{c}{New Weather} \\
  Task && Metric && CAL  & MT  & CIL & CILRS 10 & CILRS 100  && CAL  & MT  & CIL & CILRS 10 & CILRS 100   \\
  \midrule 
      \multirow{5}{*}{Empty Town } && Success &&  $84.00$ & $81.00$ & $79.00$ & $\textbf{97.33}$ & $96.33$ && $86.00$ & $85.33$ & $59.00$ & $\textbf{98.67}$ & $\textbf{98.67}$ \\ 
                                   && Col. Pedestrian &&  $0.00$ & $0.00$ & $0.00$ & $0.00$ & $0.00$ && $0.00$ & $0.00$ & $0.00$ & $0.00$ & $0.00$ \\
                                   && Col. Vehicles &&  $0.00$ & $0.00$ & $0.00$ & $0.00$ & $0.00$ && $0.00$ & $0.00$ & $0.00$ & $0.00$ & $0.00$ \\  
                                   && Col. Other &&  $9.00$ & $\textbf{11.67}$ & $11.00$ & $1.33$ & $1.33$ && $10.00$ & $9.33$ & $\textbf{33.00}$ & $0.00$ & $0.00$ \\
                                   && Timeout &&  $7.00$ & $\textbf{7.33}$ & $10.00$ & $1.33$ & $2.33$ && $4.00$ & $5.33$ & $\textbf{8.00}$ & $1.33$ & $1.33$ \\

         \midrule 
    \multirow{5}{*}{Regular Traffic} && Success &&  $57.00$ & $74.00$ & $61.50$ & $83.33$ & $\textbf{87.33}$ && $58.00$ & $68.00$ & $40.00$ & $77.33$ & $\textbf{80.00}$ \\
                                    && Col. Pedestrian &&  $7.00$ & $3.33$ & $\textbf{9.50}$ & $4.00$ & $2.00$ && $6.00$ & $8.67$ & $\textbf{15.00}$ & $2.00$ & $6.67$ \\
                                    && Col. Vehicles &&  $\textbf{26.00}$ & $6.00$ & $16.00$ & $7.67$ & $4.00$ && $\textbf{30.00}$ & $7.33$ & $17.00$ & $17.33$ & $8.67$ \\ 
                                    && Col. Other &&  $7.00$ & $11.33$ & $7.00$ & $4.67$ & $3.67$ && $2.00$ & $10.67$ & $\textbf{21.00}$ & $2.67$ & $4.00$ \\
                                    && Timeout &&  $3.00$ & $5.33$ & $6.00$ & $0.33$ & $3.00$ && $4.00$ & $5.33$ & $\textbf{7.00}$ & $0.67$ & $0.67$ \\
                                    
         \midrule
 \multirow{5}{*}{Dense Traffic}     && Success &&  $16.00$ & $\textbf{42.67}$ & $22.00$ & $\textbf{42.67}$ & $41.67$ && $18.00$ & $33.33$ & $6.00$ & $\textbf{47.33}$ & $38.00$  \\
                                    && Col. Pedestrian &&  $14.00$ & $13.67$ & $20.50$ & $\textbf{24.33}$ & $22.33$ && $12.00$ & $18.00$ & $15.00$ & $12.00$ & $14.67$ \\
                                    && Col. Vehicles &&  $\textbf{57.00}$ & $22.33$ & $49.50$ & $18.33$ & $20.67$ && $68.00$ & $18.00$ & $\textbf{69.00}$ & $26.00$ & $34.00$ \\
                                    && Col. Other &&  $10.00$ & $12.33$ & $4.50$ & $\textbf{13.33}$ & $12.33$ && $0.00$ & $\textbf{20.00}$ & $12.00$ & $11.33$ & $12.00$  \\
                                    &&Timeout &&  $3.00$ & $\textbf{9.00}$ & $3.50$ & $1.33$ & $3.00$ && $2.00$ & $\textbf{10.67}$ & $0.00$ & $3.33$ & $1.33$ \\

         \midrule
   
 \end{tabular}
}
\vspace{1mm}


\small
\centering
\resizebox{1.0\linewidth}{!}{
 \begin{tabular}{@{}lccccccccclcccccc@{}}
\toprule

                    &&  && \multicolumn{5}{c}{New Town} && \multicolumn{5}{c}{New Weather \& Town} \\
  Task && Metric && CAL  & MT  & CIL & CILRS 10 & CILRS 100  && CAL  & MT  & CIL & CILRS 10 & CILRS 100   \\
  \midrule 
      \multirow{5}{*}{Empty Town } && Success && $48.67$ & $36.33$ & $41.67$ & $66.00$ & $\textbf{72.33}$ && $57.33$ & $25.33$ & $24.00$ & $\textbf{90.67}$ & $55.33$ \\
                                   &&  Col. Pedestrian &&  $0.00$ & $0.00$ & $0.00$ & $0.00$ & $0.00$ &&  $0.00$ & $0.00$ & $0.00$ & $0.00$ & $0.00$ \\
                                   &&  Col. Vehicles &&  $0.00$ & $0.00$ & $0.00$ & $0.00$ & $0.00$ && $0.00$ & $0.00$ & $0.00$ & $0.00$ & $0.00$ \\
                                   && Col. Other &&  $45.33$ & $\textbf{56.67}$ & $51.00$ & $21.33$ & $20.00$ && $33.33$ & $64.00$ & $\textbf{66.67}$ & $5.33$ & $2.67$ \\
                                   && Timeout &&  $6.00$ & $7.00$ & $7.33$ & $\textbf{12.67}$ & $7.67$ && $9.33$ & $10.67$ & $9.33$ & $4.00$ & $\textbf{42.00}$  \\

         \midrule 
    \multirow{5}{*}{Regular Traffic} && Success &&  $27.67$ & $26.00$ & $22.00$ & $\textbf{49.67}$ & $49.00$ && $32.00$ & $14.67$ & $13.33$ & $\textbf{56.67}$ & $42.67$ \\
                                     && Col. Pedestrian &&  $6.00$ & $3.33$ & $4.33$ & $\textbf{8.33}$ & $4.33$ &&  $\textbf{7.33}$ & $2.00$ & $1.33$ & $6.67$ & $2.00$ \\
                                     && Col. Vehicles &&  $30.00$ & $9.00$ & $\textbf{34.67}$ & $8.33$ & $12.67$ && $32.67$ & $11.33$ & $\textbf{36.67}$ & $22.67$ & $18.67$ \\
                                     && Col. Other &&  $30.33$ & $\textbf{51.33}$ & $33.00$ & $21.67$ & $23.67$ && $22.67$ & $\textbf{65.33}$ & $44.00$ & $8.00$ & $8.67$ \\
                                     && Timeout &&  $6.00$ & $10.33$ & $6.00$ & $\textbf{12.00}$ & $10.33$ && $5.33$ & $6.67$ & $4.67$ & $6.00$ & $\textbf{28.00}$ \\

         \midrule
 \multirow{5}{*}{Dense Traffic}     && Success &&  $10.67$ & $9.00$ & $7.33$ & $\textbf{23.00}$ & $21.00$ && $14.67$ & $10.67$ & $2.67$ & $\textbf{24.67}$ & $12.67$ \\
                                    && Col. Pedestrian &&  $7.00$ & $8.33$ & $9.67$ & $\textbf{15.33}$ & $12.33$ && $8.00$ & $3.33$ & $4.67$ & $\textbf{14.00}$ & $8.67$ \\
                                    && Col. Vehicles &&  $46.33$ & $27.67$ & $\textbf{55.67}$ & $39.00$ & $35.00$ && $46.67$ & $38.00$ & $\textbf{57.33}$ & $37.33$ & $34.00$ \\
                                    && Col. Other &&  $28.33$ & $\textbf{40.33}$ & $24.67$ & $17.33$ & $22.67$ && $20.00$ & $\textbf{35.33}$ & $31.33$ & $7.33$ & $7.33$ \\
                                    && Timeout &&  $7.67$ & $\textbf{14.67}$ & $2.67$ & $5.33$ & $9.00$ && $10.67$ & $12.67$ & $4.00$ & $16.67$ & $\textbf{37.33}$ \\

         \midrule
   
\bottomrule
 \end{tabular}
}
\vspace{1mm}
\caption{ Analysis of the causes of episode termination for different methods. We show the results for all tasks and weather
conditions. The columns for a single method/task/condition 
should add up to 1. For each cause of episode termination we highlight
the method with higher probability (i.e. \textbf{worse} performance). For the success row, we highlight the method with the \textbf{best} performance. The reported results are
the average over three different runs of the benchmark.}
\label{tbl:cause_of_failure}
\end{table*}

}





\end{document}